\pdfoutput=1
\documentclass[11pt,table,dvipsnames]{article}

\usepackage{acl}
\usepackage{times}
\usepackage{latexsym}

\usepackage[T1]{fontenc}
\usepackage[utf8]{inputenc}

\usepackage{microtype}

\usepackage{inconsolata}
\usepackage{microtype}
\usepackage{latexsym}
\usepackage{dsfont}
\usepackage{arydshln}
\usepackage{booktabs} % For formal tables
\usepackage{subfigure}
\usepackage{CJKutf8}
\usepackage{graphicx}
\usepackage{bigstrut}
\usepackage{multirow}
\usepackage{amsmath}
\usepackage{multirow, makecell, caption}
\usepackage{floatrow}
\newfloatcommand{capbtabbox}{table}[][\FBwidth]
\usepackage{xcolor}
\usepackage[normalem]{ulem}
\usepackage{amssymb}
\usepackage{amsfonts}
\usepackage{multicol}
\usepackage{tipa}
\usepackage{amsmath}
\usepackage{bm}
\usepackage[ruled,linesnumbered]{algorithm2e}
\usepackage{tikz}
\usepackage{paralist}
\usepackage{IEEEtrantools}
\usepackage[switch]{lineno}
\usepackage{enumitem}
\usepackage{multirow, makecell, caption}
\usepackage{verbatim}
\usepackage[normalem]{ulem}
\usepackage{amssymb}
\usepackage{amsfonts}
\usepackage{multicol}
\usepackage{amssymb}% http://ctan.org/pkg/amssymb
\usepackage{pifont}% http://ctan.org/pkg/pifont
\usepackage{csquotes}
\usepackage{xspace}
\usepackage{paralist}
\usepackage{mdwlist}
\usepackage{subfigure}
\usepackage{makecell}
\usepackage{array}
\usepackage{bm}
\usepackage{colortbl}
\usepackage{dsfont}
\usepackage{url}
\usepackage{booktabs} % For formal tables
\usepackage{graphicx}
\usepackage{tabularx}
\usepackage{amsmath}
\usepackage{bbm}
\usepackage{pgfplots}
\usepackage{soul} % 导入 soul 包
\usepackage{color,xcolor} % 颜色包，color 必须导入，xcolor 建议导入
\usepackage{caption}
\usepackage{subfigure}
\usepackage{seqsplit}

\newcommand{\ie}{\textit{i.e.}\xspace}
\newcommand{\eg}{\textit{e.g.}\xspace}

\makeatletter
\def\adl@drawiv#1#2#3{%
        \hskip.5\tabcolsep
        \xleaders#3{#2.5\@tempdimb #1{1}#2.5\@tempdimb}%
                #2\z@ plus1fil minus1fil\relax
        \hskip.5\tabcolsep}
\newcommand{\cdashlinelr}[1]{%
  \noalign{\vskip\aboverulesep
           \global\let\@dashdrawstore\adl@draw
           \global\let\adl@draw\adl@drawiv}
  \cdashline{#1}
  \noalign{\global\let\adl@draw\@dashdrawstore
           \vskip\belowrulesep}}
\makeatother

\newcolumntype{m}{>{\columncolor{RoyalBlue!10}}c}
\newcolumntype{t}{>{\columncolor{RoyalPurple!10}}c}

\title{``A good pun is its own \textit{reword}'': \\ Can Large Language Models Understand Puns?}

\author{Zhijun Xu\textsuperscript{\rm $\spadesuit$},
Siyu Yuan\textsuperscript{\rm $\spadesuit$},
Lingjie Chen\textsuperscript{\rm $\spadesuit$}, 
\textbf{Deqing Yang}\textsuperscript{\rm $\spadesuit$}\thanks{~Corresponding author.}\\
\textsuperscript{\rm $\spadesuit$}School of Data Science, Fudan University\\
\texttt{\{zjxu23,syyuan21,ljchen21\}@m.fudan.edu.cn}\\ \texttt{yangdeqing@fudan.edu.cn}
}

\begin{document}

\maketitle

\begin{abstract}
As one of the common rhetorical devices, puns play a vital role in linguistic study, including the comprehensive analysis of linguistic humor. 
Although large language models (LLMs) have been widely explored on various tasks of natural language understanding and generation, their ability to understand puns has not been systematically studied, limiting the utilization of LLMs in creative writing and humor creation.
In this paper, we leverage three popular tasks, \ie, \emph{pun recognition, pun explanation}, and \emph{pun generation}, to systematically evaluate LLMs' capability of understanding puns.
In addition to the evaluation metrics adopted by prior research, we introduce some new evaluation methods and metrics that are better suited to the in-context learning paradigm of LLMs. 
These new metrics offer a more rigorous assessment of an LLM's capability to understand puns and align more closely with human cognition. 
Our research findings reveal the ``lazy pun generation'' pattern and identify the primary challenges in understanding puns with LLMs. 
The resources of this paper will be released upon publication.
%Our code is available at \url{https://github.com/Zhijun-Xu/PunEval}.
\end{abstract}

\section{Introduction}
\label{sec:introduction}
Pun, as a form of wordplay, cleverly exploits double or multiple meanings of words~\cite{miller-etal-2017-semeval}.
For example, for a pun sentence, ``A good pun is its own \underline{reword}'', it plays on the similar sounds of ``reword'' and ``reward'', suggesting that the intrinsic value or reward of a good pun lies in its clever use of language or its inventive rephrasing.
In most cases, the use of puns can produce humorous effects, as it creates a lexical-semantic ambiguity~\cite{kao2016computational} and a context-shift surprise~\cite{he-etal-2019-pun}.
Compared to other forms of humor, such as jokes~\cite{dynel2009beyond} and comedies~\cite{stott2014comedy}, puns are appropriate for linguistic humor study as they have a more precise definition and a relatively fixed structure~\cite{hempelmann2008computational,attardo2018universals}.

\begin{figure}[t]
    \centering
    \includegraphics[width=0.95\linewidth]{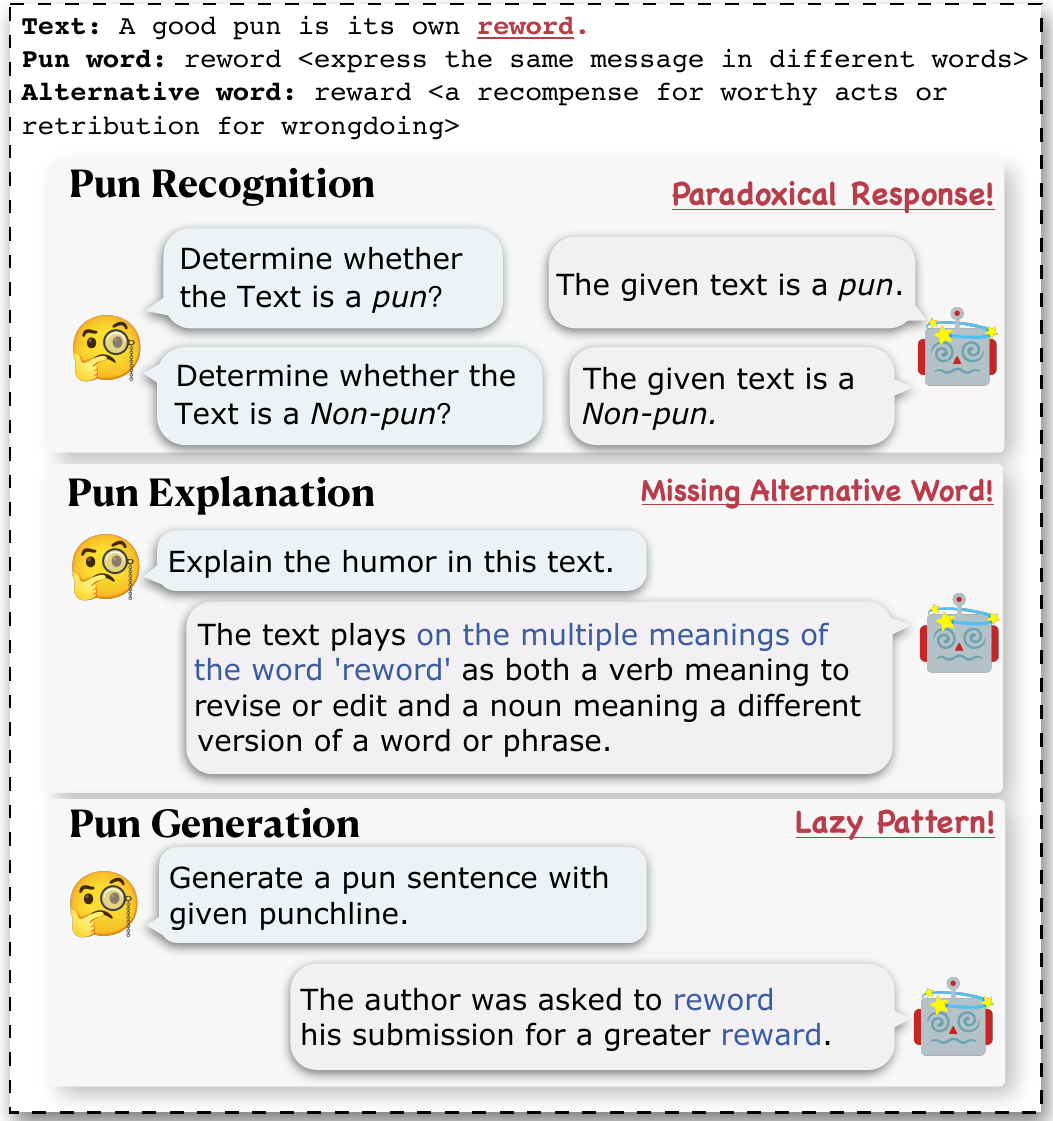}
    \caption{Toy examples of achieving three representative tasks related to pun understanding with LLMs, including pun recognition, explanation and generation. We explore the primary difficulties (\eg, paradoxical response, missing alternative word and lazy pattern) in these tasks.}
    \label{fig:011intro_image}
\end{figure}

Previous research on pun exploration primarily concentrated on developing specific language models or complex frameworks to recognize~\cite{zou-lu-2019-joint,zhou-etal-2020-boating}, explain~\cite{sun-etal-2022-expunations}, or generate~\cite{mittal-etal-2022-ambipun,tian-etal-2022-unified} puns.
With the advancement of large language models (LLMs), recent studies have explored using LLMs for detecting jokes~\cite{gupta-etal-2021-humor,baranov-etal-2023-told} and identifying humor in images\cite{hessel-etal-2023-androids} and videos~\cite{ko-etal-2023-language}.
Exploring LLMs' comprehension of puns could further enhance their values on creative text creation and humor generation.
Unfortunately, there are still no studies evaluating LLMs' capability of understanding puns systematically. 
Therefore, in this paper, we aim to systematically evaluate the capabilities of LLMs on pun understanding.
As illustrated in Figure~\ref{fig:011intro_image}, to provide comprehensive assessments, we focus on three tasks from previous work, \ie, pun recognition, pun explanation, and pun generation.
To adapt these tasks to the in-context learning (ICL) paradigm of LLMs, we develop some new methods and metrics to ensure rigorous evaluation.
For pun recognition, we create dual-biased prompts to gauge the confidence level of LLMs' responses. 
These prompts explicitly incorporate the terms "pun" or "non-pun" to interfere with the model's judgment.
For pun explanation, we employ both a fine-grained punchline check and a coarse-grained pairwise comparison. 
These methods help identify LLMs' shortcomings and assess the overall quality of LLMs' explanations. %\revise{\st{which are better aligned with human recognition.}which save labor costs and complement each other}.
For pun generation, we introduce two novel settings, \ie free and constrained generation, which demonstrate the LLMs' ability to create puns under varying conditions. 
Moreover, we introduce an \textit{Overlap} metric to measure the originality of the puns generated by LLMs.

Our research has demonstrated that most LLMs are easily influenced by prompt bias in recognizing puns.
They also struggle to explain puns which are based on phonetic similarities.
In addition, we observe that LLMs often resort to a low-quality and incorrect pattern in pun generation, separating the double meanings instead of combining them.
We term this pattern as "lazy pun generation". 
Despite all these issues, some powerful LLMs still exhibit impressive performance across the three tasks.
Specifically, LLMs are competitive with humans in pun explanation and surpass the state-of-the-art models in pun generation.
The main contributions of this paper are summarized as follows:
\begin{itemize}
    \item To the best of our knowledge, our work is the first to systematically evaluate LLMs' capabilities of pun understanding.
    \item We propose several novel evaluation methods and metrics, including dual-biased prompted asking, punchline check, and overlap indicator for assessing the originality of pun generation. 
    Compared to previous work, our evaluation methods and metrics better adapt to the ICL paradigm of LLMs. %\revise{\st{and align more closely with human cognitive processes.}}
    \item Through extensive experiments with various LLMs under different pun settings, we provide a detailed and in-depth analysis of the results. Our findings highlight the primary difficulties LLMs face in pun understanding and offer insights that could benefit future research in this area.
\end{itemize}

\section{Related Work}
\label{sec:relatedwork}
\paragraph{Studies on Puns}
Puns, recognized as a significant linguistic art form, have garnered attention in AI research~\cite{xiu-etal-2017-ecnu,doogan-etal-2017-idiom,yu-etal-2018-neural},.
Previous work mainly collects various types of puns~\cite{miller-etal-2017-semeval} from literature and the Internet and proposes diverse tasks to evaluate the pun understanding capabilities of LMs.
These tasks can be divided into three categories:
\begin{inparaenum}[\it 1)]
    \item \textit{pun recognition}~\cite{diao-etal-2018-weca,zou-lu-2019-joint,zhou-etal-2020-boating}, which involves the detection of puns and localization of pun words.
    \item \textit{pun explanation}~\cite{sun-etal-2022-expunations}, which clarifies why the puns are funny by natural language explanations.
    \item \textit{pun generation}, which mainly requests small LMs to either rewrite retrieved sentences into puns~\cite{he-etal-2019-pun,yu-etal-2020-homophonic} or create puns more flexibly using acquired context words~\cite{mittal-etal-2022-ambipun,tian-etal-2022-unified,sun-etal-2022-context}.% \revise{in recent years}. 
\end{inparaenum}
For evaluation metrics, some work analyses pun from multiple quantifiable dimensions like ambiguity and distinctiveness~\cite{kao2016computational}, as well as surprise and unusualness~\cite{he-etal-2019-pun}.
However, these studies mostly focus on training small models in pun tasks.
Our research is the first to systematically evaluate the capabilities of LLMs to recognize, explain, and generate puns.

\paragraph{LLMs for Humors}
With vastly improved understanding and creativity, LLMs not only excel in traditional humor tasks such as detection and rating~\cite{gupta-etal-2021-humor,baranov-etal-2023-told,choi-etal-2023-llms} but also demonstrate exciting potential in humor explanation and generation~\cite{jentzsch-kersting-2023-chatgpt,zhong2023let}. 
Some works aid LLMs in joke generation with humor algorithms~\cite{toplyn2023witscript} or feedback-driven techniques~\cite{ravi2024small}, while others focus on comprehending and explaining punchlines in images~\cite{hessel-etal-2023-androids} or videos~\cite{ko-etal-2023-language}.
Our work is the first to focus on pun understanding, a vital part of the humor.

\section{Preliminaries}
\label{sec:pre}
In this paper, we focus on two primary types of puns: \textit{homographic pun} (hom-pun) and \textit{heterographic pun} (het-pun)~\cite{miller-etal-2017-semeval}.
\begin{itemize}[noitemsep, leftmargin=*]
\item \textbf{Hom-Pun}: 
    Hom-puns play on the dual meaning of homographs~\cite{attardo2009linguistic}, referring to the words that have different meanings but share the same spelling.
    For example, the hom-pun ``\textit{\underline{Pick (Pick)} your friends, but not to pieces}'' utilizes the dual entendre of the word ``\textit{pick}''.
    The first part ``\textit{Pick your friends}'' suggests choosing or selecting friends. 
    However, combined with the second part ``\textit{but not to pieces}'', it evokes the phrase ``\textit{pick someone to pieces}'', meaning to criticize or find fault with someone. 
    This pun leads to an unexpected twist and creates humor.
\item \textbf{Het-Pun}:
    Het-puns leverage the double meaning of paronyms or homophones~\cite{attardo2009linguistic}, both of which are similar-sounding words but with different meanings. 
    Take the het-pun ``\textit{Life is a puzzle, look here for the missing \underline{peace (piece)}}'' as an example. 
    The word ``\textit{peace}'' typically refers to tranquility or serenity in life.
    Meanwhile, %it can easily remind one of 
    it can be easily recognized as the homophone ``\textit{piece}'', as in a puzzle piece.
    This play on ``\textit{peace}'' and ``\textit{piece}'' delivers a humorous dual entendre.
\end{itemize}
In the above two examples, the \underline{underlined} parts represent the pun-alternative word pair~\cite{he-etal-2019-pun}, with the alternative word in (parentheses).
For hom-puns, the pun word $w_p$ and the alternative word $w_a$ are identical. For het-puns, these two words have a similar pronunciation, but only the former appears in the sentence.
Both $w_p$ and $w_a$ have their respective meanings: pun sense $S_p$ and alternative sense $S_a$, which are supported by the clever use of contextual words $C_{w}$.
In the first instance, the $C_{w}$ are ``\textit{friend}'' and ``\textit{to piece}'', and in the second example, are ``\textit{life}'' and ``\textit{puzzle}''. 
Following the notation of \citet{sun-etal-2022-context}, we refer to $w_p$, $w_a$, $S_p$, and $S_a$ together as the \textit{pun pair}, denoted as $P_p=<w_p, w_a, S_p, S_a>$.

\section{Probing Protocol}
\label{sec:method}
In this section, we design an evaluation protocol consisting of three progressive tasks to assess whether LLMs can understand puns well.

\subsection{Task Formulation}

\paragraph{Task 1: Pun Recognition}
\label{subsec:methodrec}
This task requires the LLM to determine the corresponding category $C\in\{\text{pun},\text{non-pun}\}$ for a given text ${T}$, as shown in the following two examples.
\begin{quote}

\small
\textbf{Input Text}: Pick your friends, but not to pieces. \\
\textbf{Model Output}: The given text is a pun. \\
\textbf{Input Text}: A man's home is his castle. \\
\textbf{Model Output}: The given text is a non-pun.
\end{quote}

\paragraph{Task 2: Pun Explanation}
\label{subsec:methodexpl} 
This task asks the LLM to provide a natural language explanation ${E}$ for a given pun text ${T_p}$, by explicitly clarifying each element of the pun pair and the humor they express.
Here is an example:
\begin{quote}
\small
\textbf{Input Text}: Life is a puzzle, look here for the missing peace. \\
\textbf{Model Output}: The text uses the homophones "piece" and "peace". "Piece" is expected in a puzzle context, but "peace" is used, shifting the meaning to tranquility. Thus it delivers a sense of humor.
\end{quote}

\paragraph{Task 3: Pun Generation}
\label{subsec:methodgen}
This task requires the LLM to generate a pun text ${T_p}$ based on the input. 
We explore two types of inputs in our settings.
%\revise{\st{One provides only the pun pair ${P_p}$, and the other provides the pun pair ${P_p}$ along with relevant contextual words ${C_w}$.} 
Both types accept a pun pair ${P_p}$ as the basic input, but one can freely use context, while the other must utilize the given contextual words ${C_w}$.
In the following two examples, senses $S_p$ and $S_a$ are enclosed with ``<>'':
\begin{quote}
\small
\textbf{Pun Pair ${P_p}$}: peace <freedom from disputes>;\\
\hspace*{1.75cm} piece <separate part of a whole>\\
\textbf{Model Output}: When the pie was divided, everyone had a peace.\\
\textbf{Pun Pair ${P_p}$}: peace <freedom from disputes>;\\
\hspace*{1.75cm} piece <separate part of a whole>\\
\textbf{Contextual Words ${C_w}$}: life, puzzle\\
\textbf{Model Output}: In the puzzle of life, finding peace is difficult.
\end{quote}

\subsection{Task Implementation}
We design specific prompts for LLMs to test their inherent abilities on these three tasks.\footnote{All prompts for three pun-related tasks are available at Appendix~\ref{appendix:prompt}.}
\begin{itemize}[noitemsep, leftmargin=*]
    \item For \textbf{pun recognition}, we focus on the model's accuracy and confidence in its response. 
    Therefore, we craft two slightly biased instructions (one leaning towards pun and the other non-pun) in the prompt.
    We also incorporate the definition of puns and several examples into the prompt to assess their impact.
    \item For \textbf{pun explanation}, we introduce the Chain-of-Thought (CoT) technique~\cite{wei2022chain} in the recognition prompt, which requires the LLM to provide the reason before making a decision.
    The ``reason'' part is directly collected as the corresponding explanation.
    \item For \textbf{pun generation}, we employ two prompts with different requirements.
    In the free mode, LLM can freely choose its context based on the given ${P_p}$.
    In the restricted mode, LLM needs to leverage the words from \(C_w\) as much as possible.
    This enables us to evaluate the LLM's capacity to generate puns freely and under constraints.
\end{itemize}

\label{sec:dataset}
\subsection{Dataset Construction}
The dataset used in our evaluations integrates the Semeval-2017-Task-7 dataset~\cite{miller-etal-2017-semeval} with the ExPun dataset~\cite{sun-etal-2022-expunations}.
The former is a widely used open-source pun dataset, while the latter augments the former with detailed crowdsourced annotations.
Since these two datasets are not perfectly aligned and some data in ExPun lack explanations for puns, we conduct a review and filtered out some of the data.
Through this process, we ensure that each pun entry includes the pun text, pun pair, human explanation, and keyword set, whereas each non-pun entry contains only non-pun text.\footnote{We selected the longest explanation and the most extensive set of keywords in ExPun, expecting them to be more informative.}
The keyword set here serves as the contextual words $C_w$ for generating puns since it usually provides a proper context without hindering the model's generation. 
We divide the entire dataset into two parts: the hom-dataset and the het-dataset, and select a small number of samples as the demonstration examples in prompts, as shown in Table~\ref{tab:datasplit}.

\setlength\tabcolsep{4pt}
\begin{table}[t]
  \centering
  \small
    \begin{tabular}{lcccccc}
    \toprule
    \multicolumn{1}{l}{\multirow{2}[3]{*}{\textbf{Data Split}}} & \multicolumn{3}{c}{\textbf{Examples}} & \multicolumn{3}{c}{\textbf{Test Data}} \\
    \cmidrule(lr){2-4} \cmidrule(lr){5-7} &
    \multicolumn{1}{c}{hom} & \multicolumn{1}{c}{het} & \multicolumn{1}{c}{non} & \multicolumn{1}{c}{hom} & \multicolumn{1}{c}{het} & \multicolumn{1}{c}{non} \\
    \midrule
    Hom-Dataset & 10    & 0     & 10    & 810   & 0     & 633 \\
    Het-Dataset & 0     & 10    & 10    & 0     & 647   & 499 \\
    \bottomrule
    \end{tabular}%
    \caption{Dataset statistics.
    We use "hom", "het", and "non" to represent hom-puns, het-puns, and non-puns.}
  \label{tab:datasplit}%
\end{table}%

\label{sec:model}
\subsection{Model Selection}
To assess the pun understanding level of LLMs with varying parameter sizes and capabilities, we selected eight well-known LLMs from two categories for our experiments.
The first category includes open-source 7B models, such as Llama2-7B-Chat~\cite{touvron2023llama}, Mistral-7B~\cite{jiang2023mistral}, Vicuna-7B~\cite{zheng2024judging}, and OpenChat-7B~\cite{wang2024openchat}.
The second category consists of closed-source models with larger parameter scales, like Gemini-Pro~\cite{team2023gemini}, GPT-3.5-Turbo~\cite{openai2023gpt35turbo}, Claude-3-Opus~\cite{anthropic2024claude3} and GPT-4-Turbo~\cite{openai2023gpt4turbo}.
All of them are generative text models endowing with in-context learning and instruction-following abilities.

\label{sec:metrics}
\subsection{Evaluation Metrics}

\paragraph{Metrics for Recognition}
We measure the accuracy and confidence of LLMs on pun recognition through the following three indicators.
\begin{inparaenum}[\itshape 1)]
\item \textbf{True Positive Rate (TPR)}~\cite{TPRJacob1947} indicates the ratio of puns correctly identified.
\item \textbf{True Negative Rate (TNR)} is the ratio of non-puns accurately recognized.
\item \textbf{Cohen's Kappa ($\kappa$)}~\cite{cohen1960coefficient} measures the agreement between two sets of biased recognitions.
\end{inparaenum}
Moreover, we compute the \textbf{variations ($\Delta$)} in TPR and TNR when the prompt leans towards non-pun compared to pun, as they reflect the model's inconsistency intuitively.

\paragraph{Metrics for Explanation}

\setlength\tabcolsep{6pt}
\definecolor{lightgray}{gray}{0.95}
\begin{table*}[t]
  \centering
  \small
    \begin{tabular}{lmmmmmttttt}
    \toprule
    \multicolumn{1}{l}{\multirow{2}[3]{*}{\textbf{Model}}} & \multicolumn{5}{c}{\textbf{Homographic Pun}} & \multicolumn{5}{c}{\textbf{Heterographic Pun}} \\
     \cmidrule(lr){2-6} \cmidrule(lr){7-11}
    \rowcolor{white} \multicolumn{1}{c}{} &  TPR  &  $\Delta_\texttt{TPR}$  & TNR 
 &  $\Delta_\texttt{TNR}$ & \multicolumn{1}{c}{\textbf{$\kappa$}} & TPR  & $\Delta_\texttt{TPR}$ & TNR  &   $\Delta_\texttt{TNR}$ & \multicolumn{1}{c}{\textbf{$\kappa$}} \\
    \midrule
    \rowcolor{white} \multicolumn{11}{c}{\textit{Basic Prompt (with only Instruction and Test Data)}} \\
    \midrule
    Llama2-7B-Chat & \underline{0.993} & -0.128 & 0.049 & +0.294 & 0.148 & \underline{0.985} & -0.083 & 0.042 & +0.323 & 0.173 \\
    Vicuna-7B & 0.984 & -0.299 & 0.028 & +0.376 & 0.077 & \textbf{0.997} & -0.195 & 0.024 & +0.419 & 0.055 \\
    Mistral-7B & 0.867 & -0.533 & 0.208 & +0.540 & 0.156 & 0.873 & -0.442 & 0.202 & +0.585 & 0.175 \\
    OpenChat-7B & 0.948 & -0.073 & 0.368 & +0.120 & 0.722 & 0.930 & -0.068 & 0.379 & +0.120 & 0.742 \\
    \cdashlinelr{1-11}
    Gemini-Pro & \textbf{0.998} & -0.048 & 0.166 & +0.506 & 0.287 & 0.983 & -0.133 & 0.192 & +0.467 & 0.296 \\
    GPT-3.5-Turbo & 0.990 & -0.137 & 0.224 & +0.510 & 0.291 & 0.977 & -0.148 & 0.263 & +0.467 & 0.342 \\
    Claude-3-Opus & 0.989 & \underline{-0.011} & \underline{0.624} & \underline{+0.109} & \underline{0.867} & 0.969 & \underline{-0.037} & \underline{0.613} & \underline{+0.096} & \underline{0.839} \\
    GPT-4-Turbo & 0.988 &  \textbf{-0.003} & \textbf{0.630} & \textbf{+0.054} & \textbf{0.894} & 0.960 & \textbf{-0.020} & \textbf{0.621} & \textbf{+0.048} & \textbf{\text{0.884}} \\

    \midrule
    \rowcolor{white} \multicolumn{11}{c}{\textit{Enhanced Prompt (with Additional Pun Definition and 6 Examples)}} \\
    \midrule
    Llama2-7B-Chat & 0.738 & +0.123 & 0.306 & -0.071 & 0.309 & 0.770 & +0.153 & 0.501 & -0.313 & 0.208 \\
    Vicuna-7B & 0.986 & \textbf{-0.001} & 0.112 & \underline{+0.016} & 0.726 & \underline{0.985} & \textbf{+0.000} & 0.283 & \underline{+0.044} & 0.842 \\
    Mistral-7B & 0.569 & -0.181 & \underline{0.798} & +0.076 & 0.696 & 0.553 & -0.158 & \textbf{0.894} & +0.064 & 0.722 \\
    OpenChat-7B & 0.890 & -0.063 & 0.556 & +0.107 & 0.816 & 0.873 & -0.060 & 0.667 & +0.048 & 0.881 \\
    \cdashlinelr{1-11}
    Gemini-Pro & \textbf{0.998} & -0.058 & 0.460 & +0.422 & 0.519 & 0.982 & -0.097 & 0.499 & +0.349 & 0.555 \\
    GPT-3.5-Turbo & 0.974 & -0.036 & 0.611 & +0.137 & 0.811 & 0.935 & -0.056 & 0.699 & +0.106 & 0.814 \\
    Claude-3-Opus & 0.982 & \underline{-0.005} & \textbf{0.806} & +0.041 & \underline{0.953} & \textbf{0.991} & \underline{-0.003} & 0.750 & +0.070 & \underline{0.929} \\
    GPT-4-Turbo & \underline{0.988} & \textbf{-0.001} & 0.758 & \textbf{+0.010} & \textbf{0.962} & 0.961 & +0.008 & \underline{0.796} & \textbf{-0.006} & \textbf{0.959} \\

    \bottomrule
    \end{tabular}%
    \caption{Results of two biased pun recognition.
    Apart from TPR, TNR, and $\kappa$, we also compute the variations ($\Delta$) in TPR and TNR when the prompt bias shifts from pun to non-pun.
    These variations are similarly marked based on their absolute values.
    The best results (smallest variations) are \textbf{bolded}, and the second-best results are \underline{underlined}.}
  \label{tab:recognition}
\end{table*}

Considering the labor-intensive and time-consuming nature of manually evaluating pun explanation, we combine manual assessment with automatic evaluation according to the following two methods.
\begin{inparaenum}[\itshape 1)]
    \item A small-scale, fine-grained \textit{punchline check}: 
    We randomly select 100 hom-puns and 100 het-puns and employ three annotators to assess the quality of their explanations.\footnote{More information about our annotators can be found in Appendix~\ref{appendix:crowd-sourcing}.}
    For each sample, we ask annotators to check whether elements of the pun pair $P_p=<w_p, w_a, S_p, S_a>$ are correctly mentioned in the explanation.
    Their annotations demonstrate a high level of agreement (with Fleiss’s $\kappa=0.87$), highlighting the reliability of this method.
    In cases of disagreement, we adopt the majority view.
    Then, we compute the average mentioned ratio (denoted as \textbf{Average Mention Ratio}) of $w_p$, $w_a$, $S_p$, and $S_a$ as indicators.
    \item A large-scale coarse-grained \textit{pairwise comparison}: 
    We instruct GPT-4~\cite{openai2023gpt4} to choose the winner between the human explanation and the model explanation (allowing for a tie), and then calculate the \textbf{Win Rate}, \textbf{Tie Rate}, and \textbf{Loss Rate} of each LLM. This kind of approach is widely used for evaluation~\cite{li2024translate,yuan2024easytool,qin2024toolllm}.
    It is worth noting that GPT-4 achieves a high level of consistency with our annotators, showing an accuracy of 88.3\% on the sampled data. 
\end{inparaenum}

\paragraph{Metrics for Generation}
\label{subsec:genmetric}
The metrics used for pun generation in our study consist of two main dimensions:
\begin{inparaenum}[\itshape 1)]
    \item automatic indicators, which are primarily based on word probability modeling, like \textbf{Ambiguity (A)}, \textbf{Distinctiveness (D)}~\cite{kao2016computational}, and \textbf{Surprise (S)}~\cite{he-etal-2019-pun}.\footnote{The formula for calculating these metrics and the details of their implementation are available at Appendix~\ref{appendix:adsu}}.
    %\revise{We also compute \textbf{One-pun-word Incorporation Rate ($1w_{p}$)} and \textbf{Contextual Word Incorporation Rate}~\cite{sun-etal-2022-context} as a supplement.}
    We also examine the inclusion rates of the pun and the contextual word in the generation, denoted as \textbf{One-pun-word Incorporation Rate ($1w_{p}$)} and \textbf{Contextual Word Incorporation Rate}~\cite{sun-etal-2022-context}.
    \item manual indicators, which include \textbf{Success Rate} and \textbf{Funniness Rating} of human puns and LLM-generated puns.
    We ask our annotators to identify whether a pun text is successful and rate its funniness on a scale from 1 to 5. 
    These annotations are performed on the same subset (100 hom-puns and 100 het-puns) of our dataset.
\end{inparaenum}

\section{Results and Analysis}
\label{sec:results}
\subsection{\textit{Can LLMs Distinguish Between Puns and Non-puns?}}

We design two types of prompts for pun recognition: 
The first type is the \textit{basic prompt}, which only includes test data and biased instructions.
The second type is the \textit{enhanced prompt}, which adds to the basic prompt with the definition of puns and some examples (3 puns and 3 non-puns).

\begin{figure}[t]
    \centering
    \includegraphics[width=0.98\linewidth]{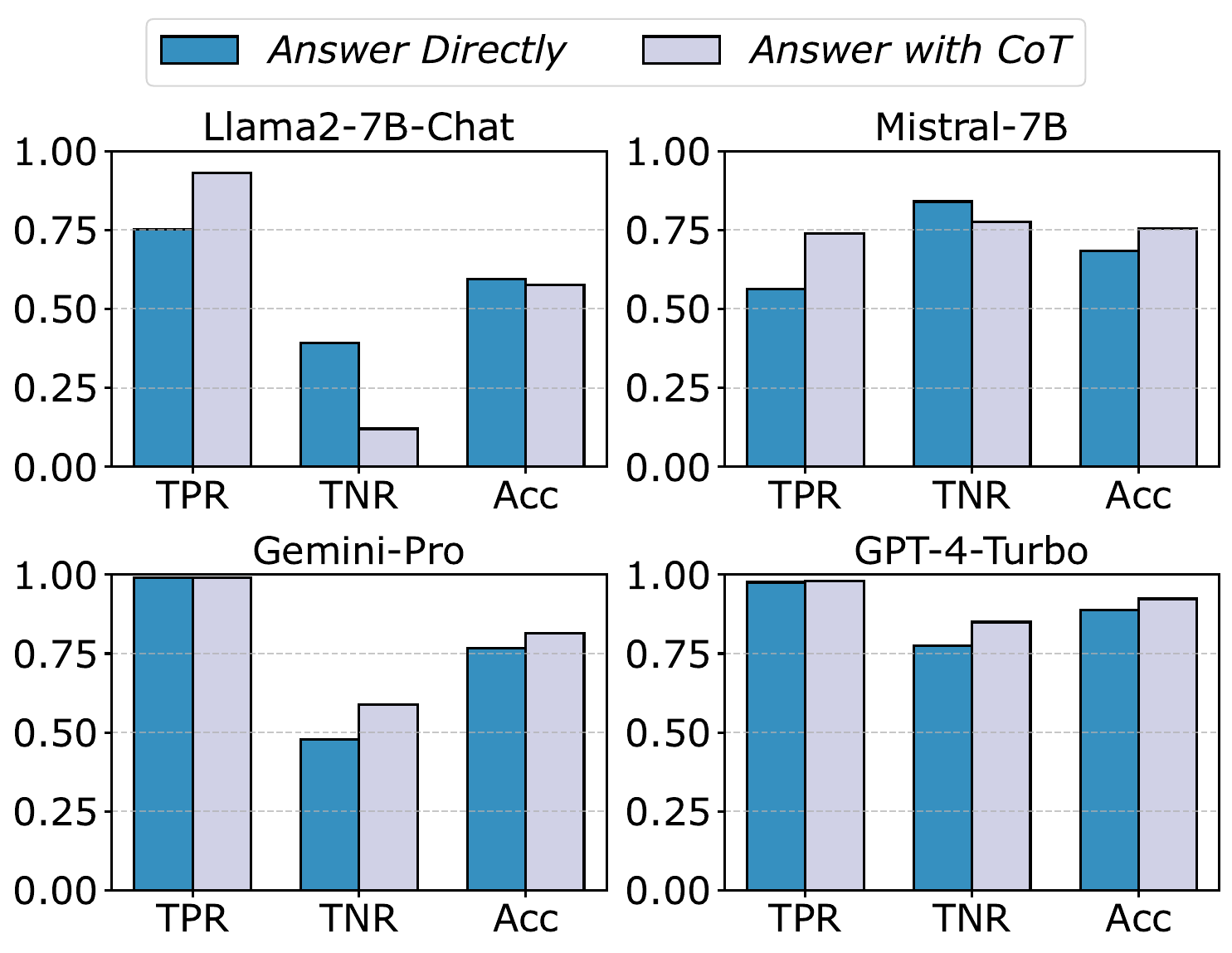}
    \caption{The performance of four selected LLMs in recognizing puns via direct answers and CoT responses.
    The Acc metric represents the overall accuracy.}
    \label{fig:recog_cot}
\end{figure}

As shown in Table~\ref{tab:recognition}, we can find that:
\begin{inparaenum}[\itshape 1)]
    \item Almost all tested LLMs are influenced by the bias in the prompt, leading to results that tend to align with this bias.
    Some models, such as Vicuna-7B, Mistral-7B, Gemini-Pro, and GPT-3.5-Turbo, show significant fluctuations in their responses, indicating their lack of confidence in their answers.
    \item Adding a definition and examples as additional information significantly improves the consistency between LLMs' two responses.
    It also enhances the models' accuracy in recognizing non-puns.
    \item The TNR metric is generally lower than the TPR.
    This discrepancy arises because non-puns in our dataset are mostly non-pun jokes and proverbs. They are somewhat similar to puns.
    \item There is no obvious difference in recognizing hom-puns and het-puns. This may reveal that LLMs capture the core feature (\ie, dual meanings) of puns and use it as the main criterion for judgment.
    \item GPT-4-Turbo and Claude-3-Opus demonstrate exceptional performance, exhibiting satisfactory pun recognition capabilities.
\end{inparaenum}

\setlength\tabcolsep{2pt}
\begin{table}[t]
  \centering
  \small
    \begin{tabular}{lmmmmtttt}
    \toprule
    \multirow{2}[3]{*}{\textbf{Model}} & \multicolumn{4}{c}{\textbf{Homographic Pun}} & \multicolumn{4}{c}{\textbf{Heterographic Pun}}  \\
    \cmidrule(lr){2-5} \cmidrule(lr){6-9} 
     \rowcolor{white} \multicolumn{1}{c}{} & \textbf{$w_p$} & \textbf{$w_a$} & \textbf{$S_p$} & \textbf{$S_a$} & \textbf{$w_p$} & \textbf{$w_a$} & \textbf{$S_p$} & \textbf{$S_a$} \\
    \midrule
    Llama2-7B-Chat & 0.63  & 0.63  & 0.45  & 0.42  & 0.69  & 0.11  & 0.47  & 0.13   \\
    Vicuna-7B & 0.71  & 0.71  & 0.64  & 0.59  & 0.85  & 0.21  & 0.81  & 0.29  \\
    Mistral-7B & 0.78  & 0.78  & 0.73  & 0.68  & 0.69  & 0.22  & 0.68  & 0.22  \\
    OpenChat-7B & 0.81  & 0.81  & 0.72  & 0.71  & 0.77  & 0.28  & 0.74  & 0.33   \\
    \cdashlinelr{1-9}
    Gemini-Pro & 0.92  & 0.92  & 0.87  & 0.81  & 0.89  & 0.42  & 0.83  & 0.42  \\
    GPT-3.5-Turbo & 0.88  & 0.88  & 0.81  & 0.81  & 0.91  & 0.55  & 0.82  & 0.57 \\
    Claude-3-Opus & \underline{0.96}  & \underline{0.96}  & \underline{0.95}  & 0.92  & 0.95  & 0.84  & \textbf{0.94} & 0.78 \\
    GPT-4-Turbo & \textbf{0.98} & \textbf{0.98} & \textbf{0.96} & \underline{0.93}  & \underline{0.96}  & \underline{0.90}   & \underline{0.93}  & \underline{0.85}  \\

    \midrule
    Human & 0.95  & 0.95  & \underline{0.95}  & \textbf{0.95} & \textbf{0.97} & \textbf{0.97} & \textbf{0.94} & \textbf{0.93} \\
    \bottomrule
    \end{tabular}%
    \caption{Results of punchline check for pun explanations. We represent the average mention ratio of the pun pair elements in explanations with the corresponding symbols. The top outcomes are \textbf{bolded} and the second best are \underline{underlined}.
    }
    \label{tab:punchlinecheck}
\end{table}%

\paragraph{CoT Prompting}
Although we primarily use CoT to obtain explanations of puns from LLMs, it also offers an opportunity to explore its impact on the pun recognition task.
We differentiate between two response methods based on the enhanced prompt: answering directly and answering with CoT, while keeping the prompt's bias towards pun. 
Then, we select four models and chart their performance in Figure~\ref{fig:recog_cot}.
It is observable that, except for LLama2-7B-Chat, the remaining three LLMs showed an overall improvement in accuracy after using CoT.
Notably, Gemini-Pro and GPT-4-Turbo's weak spots in recognizing non-pun text are compensated for through CoT response, showcasing a stronger ability to distinguish between puns and non-puns.

\subsection{\textit{Can LLMs Explain the Humor in Puns?}}

\begin{figure}[t]
    \centering
    \includegraphics[width=0.95\linewidth]{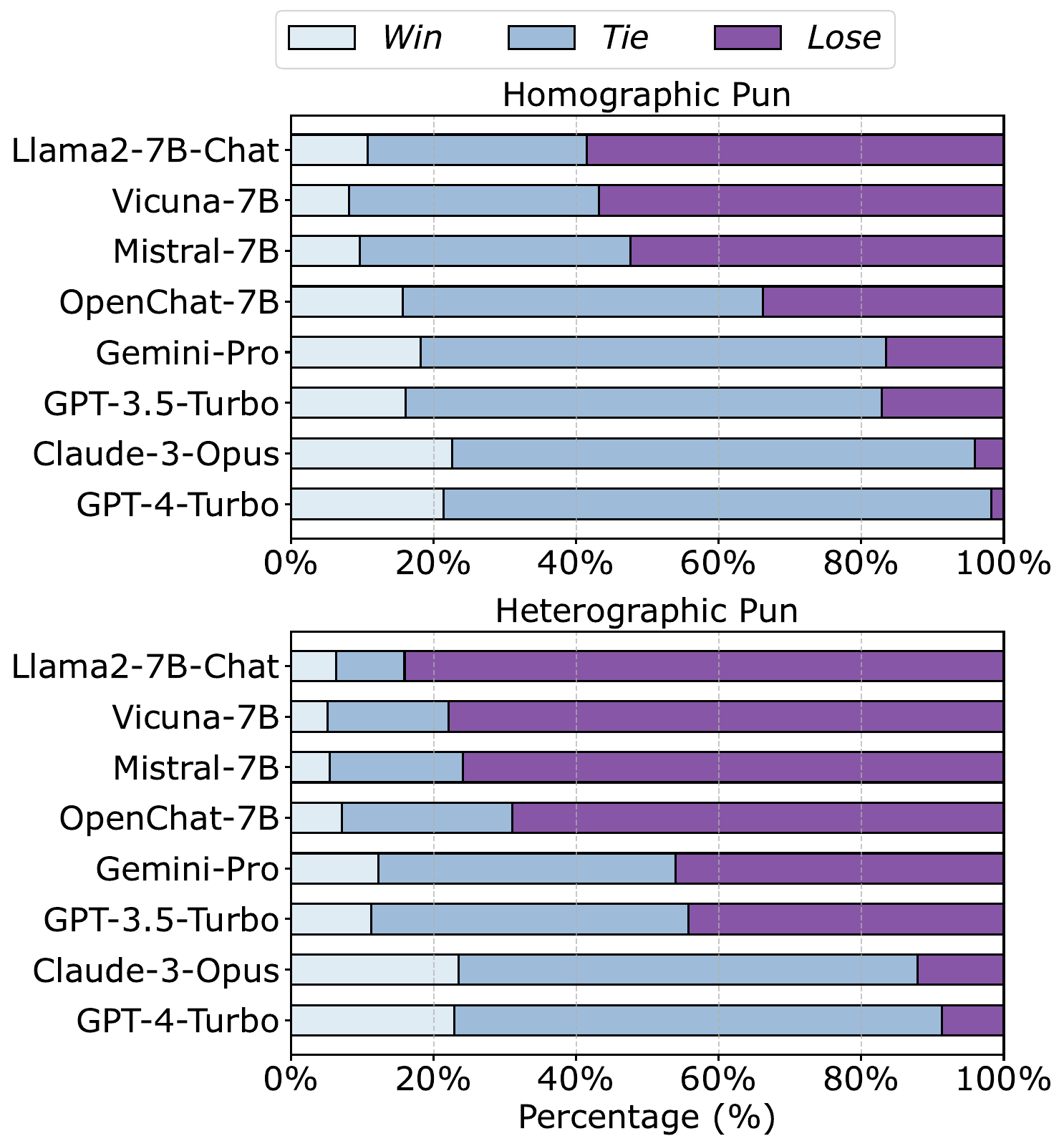}
    \caption{Results of pairwise comparison for pun explanations}
    \label{fig:pairwise}
\end{figure}

The humor in puns mainly stems from exploiting double entendre.
Thus, explaining the humor in a pun is akin to identifying its dual meanings or, more precisely, the corresponding pun pair.

We present the results of the punchline check in Table~\ref{tab:punchlinecheck}. 
This evaluation shows that:
\begin{inparaenum}[\itshape 1)]
    \item Most LLMs accurately identify the pun words $w_p$ in both hom-puns and het-puns, which is fundamental to explaining puns.
    \item Except for GPT-4-Turbo and Claude-3-Opus, the remaining LLMs struggle to identify alternative words $w_a$ and alternative sense $S_a$ in het-puns.
    This challenge arises because $w_a$ in het-puns does not directly appear in the text but relies on evocation through context and similar pronunciation to $w_p$. 
\end{inparaenum}

\definecolor{lightgray}{gray}{0.95}
\setlength\tabcolsep{3.5pt}
\begin{table*}[ht]
  \centering
  \small
    \begin{tabular}{lmmmmmmtttttt}
    \toprule
   \multirow{2}[3]{*}{\textbf{Model}} & \multicolumn{6}{c}{\textbf{Homographic Pun}}  & \multicolumn{6}{c}{\textbf{Heterographic Pun}} \\
    \cmidrule(lr){2-7}  \cmidrule(lr){8-13}
    \rowcolor{white} & {\textbf{A}}  & {\textbf{D}}   & {\textbf{S}}  & \textbf{$1w_p$} & Success &  Funny & {\textbf{A}}     & {\textbf{D}}   & {\textbf{S}}  & \textbf{$1w_p$} & Success  & Funny \\
    \midrule
    
    Generated non-pun & 0.195  & 0.037  & -0.640 & 0.983 & 0.010 & 1.042  & 0.113  & 0.071  & -0.734  & 0.978 & 0.010 & 1.014 \\
    \midrule
    \rowcolor{white} \multicolumn{13}{c}{\textit{Pun Generation with only Pun Pair}} \\
    \midrule
    Llama2-7B-Chat & 0.206  & 0.033  & \underline{-0.130}  & 0.425  & 0.060 & 1.071  & 0.168  & \underline{0.155}  & -0.029  & 0.145  & 0.040 &  1.042 \\
    Vicuna-7B & \underline{0.223}  & 0.062  & -0.249  & 0.690  & 0.120  &  1.216  & 0.211  & 0.088  & -0.272  & 0.377  & 0.050  &  1.128 \\
    Mistral-7B & 0.193  & \underline{0.072}  & -0.239  & 0.583  & 0.170  & 1.321  & 0.211  & 0.151  & -0.156  & 0.343  & 0.130 &  1.336  \\
    OpenChat3.5-7B & 0.208  & 0.058  & -0.168  & 0.549  & 0.200 & 1.261  & 0.207  & 0.136  & -0.261  & 0.271  & 0.080 &  1.128  \\
    \cdashlinelr{1-13}
    Gemini-Pro & 0.222  & 0.038  & -0.203 & 0.680  & 0.320  & 1.699 & \textbf{0.241 } & 0.072  & -0.076  & 0.383  & 0.150 &  1.336  \\
    GPT3.5-Turbo & 0.220  & 0.064  & -0.233  & 0.714  & 0.420  & 1.367  & \underline{0.223}  & 0.073  & 0.072  & 0.521  & 0.290 & 1.306  \\
    Claude3-Opus & 0.211  & \textbf{0.073 } & -0.150  &  \textbf{0.893 } & \underline{0.540}  & \underline{2.050}  & 0.200  & \textbf{0.208 } & \underline{0.096}  & \textbf{0.915 } & \underline{0.470} & \underline{1.931}  \\
    GPT4-Turbo & \textbf{0.225 } & 0.047  & \textbf{-0.027 }  & \underline{0.890}  & \textbf{0.600}  & \textbf{2.016} & 0.221  & 0.098  & \textbf{0.121 }  & \underline{0.847}  & \textbf{0.510}  & \textbf{1.948}  \\

    \midrule
    \rowcolor{white} \multicolumn{13}{c}{\textit{Pun Generation with Pun Pair and Relevant Contextual Words}} \\
    \midrule
    Llama2-7B-Chat & 0.205  & \underline{0.107}  & \underline{-0.093} & 0.605  & 0.340 & 1.602  & 0.180  & \textbf{0.235 } & -0.066  & 0.352  & 0.220 & 1.413  \\
    Vicuna-7B & 0.199  & 0.077  & -0.181  & 0.782  & 0.300 & 1.650  & 0.182  & \underline{0.238}  & 0.015  & 0.453  & 0.210 & 1.459  \\
    Mistral-7B & 0.186  & \textbf{0.115 } & -0.201  & 0.616  & 0.280 & 1.618 & 0.176  & 0.213  & 0.108  & 0.373  & 0.220 & 1.506  \\
    OpenChat3.5-7B & 0.196  & 0.091  & -0.133  & 0.636  & 0.370  &  1.715  & 0.166  & \textbf{0.235 } & 0.013  & 0.352  & 0.240 & 1.522  \\
    \cdashlinelr{1-13}
    Gemini-Pro & \underline{0.221}  & 0.079  & -0.200  & 0.689  & 0.440 & 1.880  & 0.198  & 0.149  & 0.142  &  0.581  & 0.330  & \underline{1.731}  \\
    GPT3.5-Turbo & 0.217  & 0.079  & \textbf{-0.076 }  & 0.856  & 0.550  &  2.137 & \textbf{0.216 } & 0.163  & 0.205  & 0.543  & 0.320  & 1.699   \\
    Claude3-Opus & \textbf{0.237 } & 0.081  & -0.131  &  \textbf{0.907 } & \underline{0.650} & \underline{2.438}  & \underline{0.206}  & 0.185  & \underline{0.275}  & \textbf{0.849 } &  \textbf{0.610}  & \textbf{2.348} \\
    GPT4-Turbo & 0.217  & 0.082  & -0.217 & \underline{0.880}  & \textbf{0.670} &  \textbf{2.584} & 0.199  & 0.168  & \textbf{0.285 }  & \underline{0.794}  & \underline{0.600} & \textbf{2.348}  \\

    \midrule
    Human pun & {{0.225}} & {{0.129}} & {{-0.069}}  & {{0.990}} & 0.860 & 3.268 & {{0.185}} & {{0.256}} & {{0.323}} & {{0.985}} &  0.840 & 3.229 \\
    \bottomrule
    \end{tabular}%
    \caption{Results of pun generation.
    We abbreviate the metrics Ambiguity, Distinctiveness, Surprise, and One-pun-word Incorporation Rate as "A", "D", "S" and "$1w_p$", respectively.
    For each generation method, the best results appear in \textbf{bold} and the second best are \underline{underlined}.
    }
  \label{tab:generation}%
\end{table*}%

Unlike the detail-oriented punchline check, pairwise comparison focuses on the overall quality of explanations.
Its results, illustrated in Figure~\ref{fig:pairwise}, indicate that:
\begin{inparaenum}[\itshape 1)]
    \item LLMs generally perform worse at explaining het-puns than hom-puns, aligning with the findings in the punchline check.
    Based on the results of pun recognition, we infer that alternative words do not affect pun recognition but are crucial for correctly explaining puns.
    \item The explanations by GPT-4-Turbo and Claude-3-Opus often approach or even surpass those by humans. 
    We find that LLMs consistently use a general-to-specific structure in their explanations, whereas human explanations tend to be more casual.\footnote{The structure of the pun explanation is further discussed in Appendix~\ref{appendix:structure}}
    This aspect gives the two models an edge in comparison.
\end{inparaenum}

\paragraph{Error Types in Explanation}
LLMs tend to make various mistakes when explaining puns, and we categorize the primary errors as follows:
\begin{inparaenum}[\itshape 1)]
\item \textit{Misclassify pun as non-pun}, which means the model fails to detect the double meaning.
\item \textit{Incorrect pun word identification}, which means the model fails to find the correct $w_p$.
\item \textit{Incorrect alternative word identification}, a mistake only made in the explanation of het-puns, which means the model fails to evoke the correct $w_a$.
\item \textit{Misinterpret het-pun as hom-pun}, which means the model wrongly classifies the pun's genre.
\item \textit{Lack of meaning analysis}, which means the model points out $w_p$ and $w_a$ but skips explaining the dual meanings.
\item Fabricating non-existent meanings, which means the model invents meanings for $w_p$ or $w_a$ that do not exist.
We provide a case for each type of error in Appendix~\ref{appendix:explerror} to help readers understand them.
We believe addressing these errors is key to enabling LLMs to generate better explanations of puns.
\end{inparaenum}

\subsection{\textit{Are LLMs Capable of Generating Puns?}}
To answer this question, we first ask GPT-3.5-Turbo to generate non-puns containing the same pun words $w_p$ as human puns, to serve as a baseline.
Then, we request all tested LLMs to generate puns under two different inputs mentioned in $\mathsection$~\ref{subsec:methodgen}.

\begin{figure}[t]
    \centering
    \includegraphics[width=0.95\linewidth]{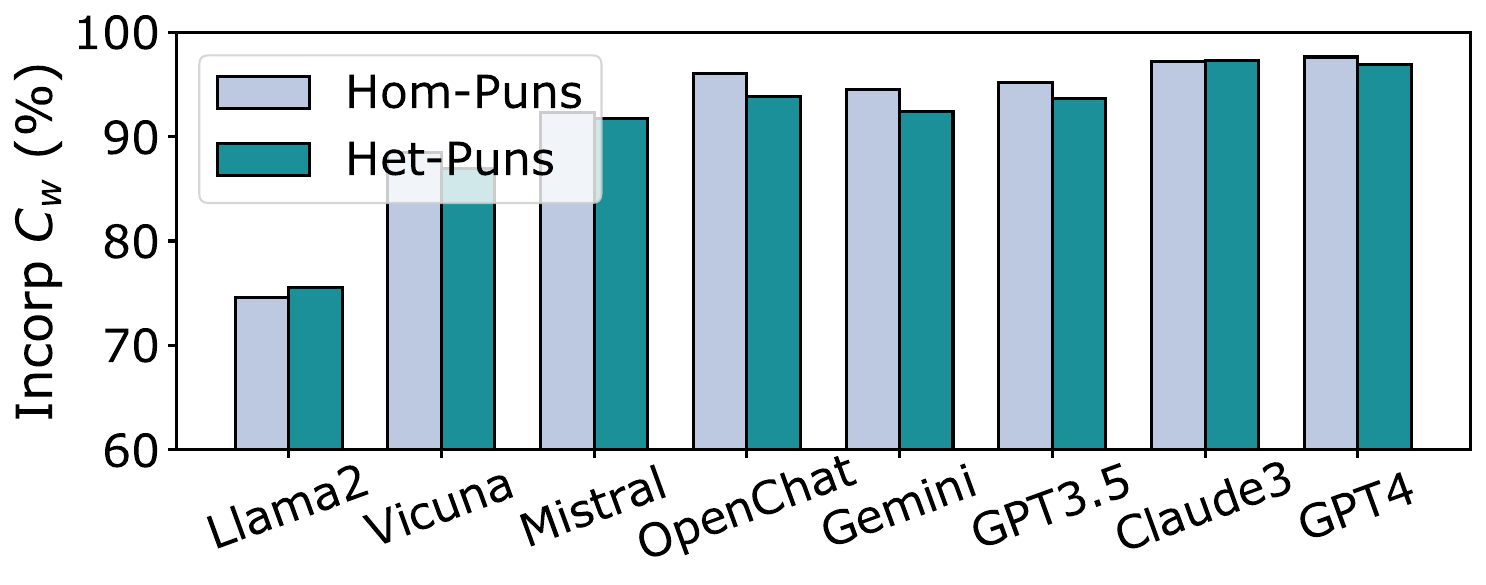}
    \caption{Contextual word incorporation rate of different LLMs in constrained pun generation}
    \label{fig:incorp_contexts}
\end{figure}

From Figure~\ref{fig:incorp_contexts}, we can see that with the exception of LLama2-7B-Chat, all other LLMs can easily accomplish the task of constrained generation.
They are notably efficient at incorporating nearly all contextual words $C_w$ in the generated sentences.
Other metrics are presented in Table~\ref{tab:generation}.
Our analysis reveals that:
\begin{inparaenum}[\itshape 1)]
    \item All LLMs demonstrate a noticeably weaker ability to generate het-puns than hom-puns, indicating that het-pun generation is a more challenging task.
    \item Since all $C_w$ are derived from human puns, we believe LLMs can grasp the intrinsic relationship between these words and the given pun pair, thereby improving the quality and success rate of the generated puns.
    This suggests that providing good context helps in pun generation.
    \item Most LLMs, especially the 7B models, tend to include multiple $w_p$  when generating puns. 
    This phenomenon is rarely seen in human puns, which usually leads to the failure of pun generation.
    \item GPT-4-Turbo and Claude-3-Opus achieve impressive success in generating puns and rival the traditional state-of-the-art methods, which has a success rate of 56\% for hom-puns and 47\% for het-puns~\cite{tian-etal-2022-unified}. However, the puns they generate are still not as funny as those created by humans.
\end{inparaenum}

\paragraph{``Lazy Pun Generation'' Pattern}

\begin{table}[t]
\footnotesize
  \centering
    \begin{tabularx}{\linewidth}{X}
    \toprule
    \rowcolor[gray]{0.95}
    \multicolumn{1}{c}{\textbf{Lazy Pun Generation Samples}} \\
    \midrule
    \makecell[{{p{\linewidth}}}]{
    \vspace{-2.5mm}
    \color{gray}{/* \textit{Pun Pair} */}\\
    dock <deprive someone of benefits, as a penalty>\\
    dock: <come into dock>\\
    \vspace{-2mm}
    \color{gray}{/* \textit{Human Pun} */}\\
    When longshoremen show up late for work they get \underline{docked}.\\
    \vspace{-2mm}
    \color{gray}{/* \textit{LLM Generation} */}\\
    The sailor's pay was \underline{docked} after he struggled to \underline{dock} on time.\\
    }\\ 
    \midrule
    \makecell[{{p{\linewidth}}}]{
    \vspace{-2.5mm}
    \color{gray}{/* \textit{Pun Pair} */}\\
    two <the cardinal number that is the sum of one and one>\\
    too <to a degree exceeding normal or proper limits>\\
    \vspace{-2mm}
    \color{gray}{/* \textit{Human Pun} */}\\
    My friend gave me a book about puns for my birthday and I loved it. It was \underline{two} meaningful.\\
    \vspace{-2mm}
    \color{gray}{/* \textit{LLM Generation} */}\\
    I tried to make puns about numbers, but \underline{two} were \underline{too} much to handle.\\
    }\\ 
    \bottomrule
    \end{tabularx}
  \caption{Examples of LLMs' lazy pun generation pattern.
  We \underline{underline} the $w_p$ and $w_a$ in human puns and LLM-generated puns.}
  \label{tab:gencase}
\end{table}

% hom_925
% The pun pair: 
% dock <deprive someone of benefits, as a penalty>
% dock <come into dock>
% Human pun
% When longshoremen show up late for work they get docked.
% Model pun
% The sailor's pay was docked after he struggled to dock on time.

% het_900
% The pun pair: 
% two <the cardinal number that is the sum of one and one>
% too <to a degree exceeding normal or proper limits>
% Human pun
% My friend gave me a book about puns for my birthday and I loved it. It was two meaningful.
% Model pun
% I tried to make puns about numbers, but two were too much to handle.

No matter how much the prompt emphasizes that only one $w_p$ should be used, most LLMs frequently generate text containing two or even more $w_p$ (and $w_a$ for het-puns), as shown in Table~\ref{tab:gencase}.
We refer to this stubborn pattern as \textit{lazy pun generation}, and classify pun sentences produced in this pattern as unsuccessful.
We attribute this pattern to two main reasons.
Firstly, including multiple $w_p$ allows for expressing double meanings at different parts of the sentence, making the construction relatively simple.
Secondly, the current definitions of puns do not explicitly limit the number of $w_p$ and $w_a$ used. 
Avoiding $w_a$ in het-puns and adopting a single $w_p$ is an unwritten rule that most human-crafted puns follow, but LLMs often ignore.
Since adding corresponding restrictions in the prompt can slightly alleviate this issue, we believe it would be more helpful for the LLM to learn this explicitly through definitions or cases during training.

\paragraph{Copying or Originality?}

\begin{figure}[t]
    \centering
    \includegraphics[width=0.95\linewidth]{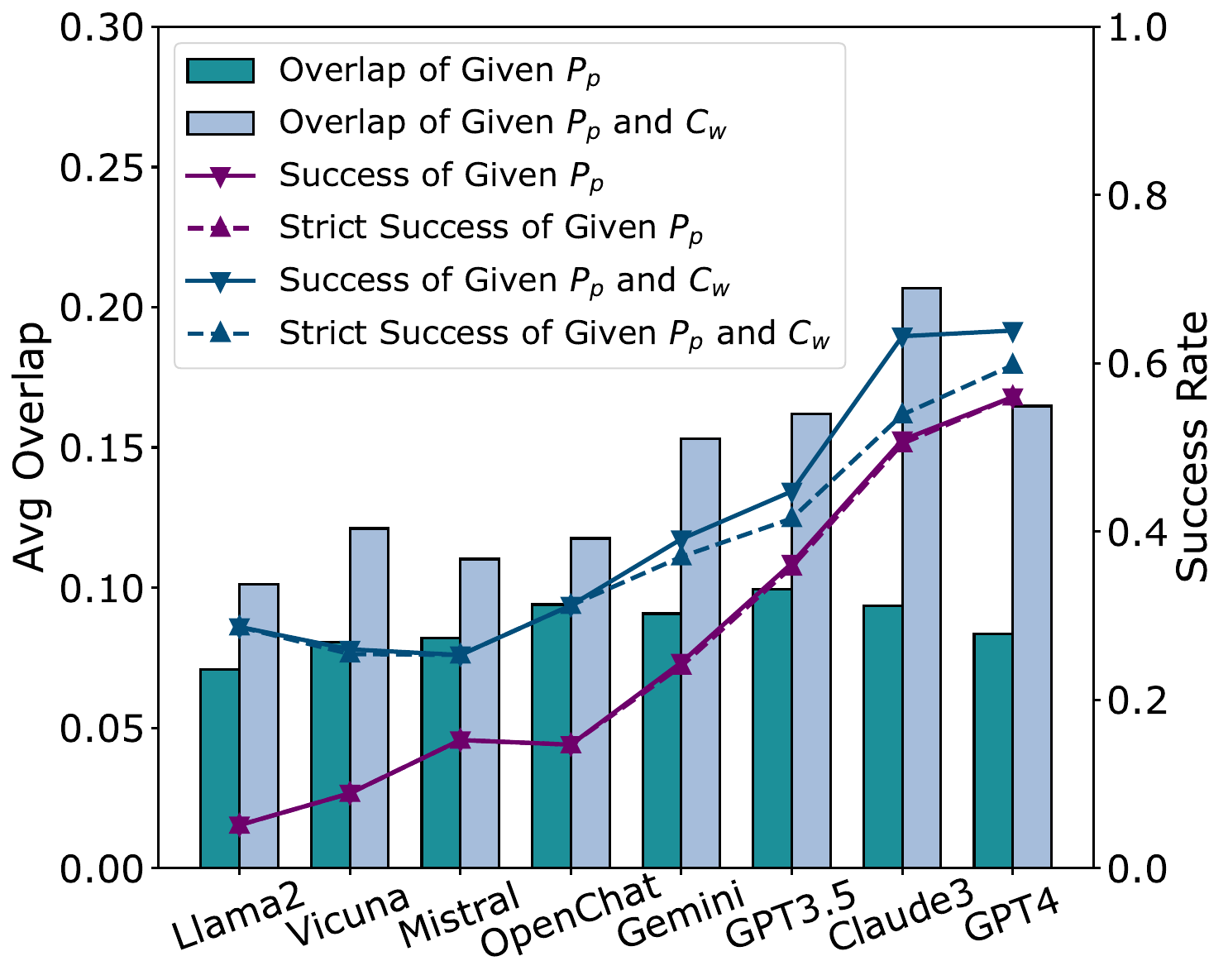}
    \caption{Average overlap, success, and strict success of two methods for generating puns.}
    \label{fig:overlap}
\end{figure}

LLMs are trained on vast amounts of text.
It's essential to ascertain whether they merely reproduce existing puns or genuinely create new ones.
To assess this, we developed an \textit{Overlap} metric to measure the similarity between puns created by models and those by humans.
The metric's computation involves three steps.
First, we identify the lemma word sets in puns generated by LLMs and humans, labeled as ${Pun}_{LLM}$ and ${Pun}_{human}$.
Next, we eliminate the words $w_p$, $w_a$, and $C_w$ provided in the prompt, resulting in refined sets $\tilde{Pun}_{LLM}$ and $\tilde{Pun}_{human}$.
Finally, we compute the overlap ratio as the size of the intersection over the size of the union of these sets, as in the formula:
$$
Overlap = \frac{\lvert{\tilde{Pun}}_{LLM} \cap {\tilde{Pun}}_{human} \rvert}
{\lvert{\tilde{Pun}}_{LLM} \cup {\tilde{Pun}}_{human} \rvert}
$$
We establish a coarse criteria for originality as an overlap < 0.5, thereby defining the ``\textbf{\textit{Strict Success}}'' of pun generation, which combines success with originality.
Figure~\ref{fig:overlap} shows that:
\begin{inparaenum}[\itshape 1)]
    \item When given only pun pair $P_p$, LLMs rarely copy human puns, relying probably on self-creation.
    \item When given additional $C_w$, the likelihood of LLMs reproducing human puns increases slightly, leading to a decrease in strict success.
    We also find that the larger the LLM, the more prone it is to do this, suggesting that their stronger memory of the corpus adversely affects the generation of creative puns.
\end{inparaenum}

\section{Conclusion}
\label{sec:conclusion}
In this paper, we examine the ability of large language models (LLMs) to understand puns. We employ three tasks: pun recognition, pun explanation, and pun generation, and develop various metrics to systematically assess the capabilities of LLMs in these areas.
Experiments indicate that although LLMs perform satisfactorily in recognizing and explaining puns, there is still room for improvement in their ability to generate creative and humorous puns. 
We also suggest that het-pun explanation and generation are more difficult than those of hom-pun.
We believe our evaluation methods and findings will contribute to advancing research on pun understanding.

\section*{Limitations}
\label{sec:limitation}
Although we utilize the most widely used pun dataset currently available to evaluate the pun-understanding ability of LLMs, our pun texts are all in English.
The ability of LLMs to understand puns can vary across different languages, and puns in languages other than English may have different definitions, structures, or purposes. 
Such a limitation highlights the potential for future work to generalize to puns in other languages.

In addition, given that LLMs have massive training data, most of which are not publicly available, it is possible that LLMs just copy puns that are not present in our dataset.
Thus, our \textit{Overlap} metric is not a precise measurement but only roughly indicates the extent of LLMs' plagiarism when generating puns.
%It is not a precise measurement.
Since exploring originality is intriguing, we eagerly hope for future work to develop more accurate indicators.

Another limitation of our work stems from potential biases in the evaluation process.
Evaluating the quality of a pun explanation and the success of a generated pun involves human annotator judgments. 
Preferences vary among individuals: some may prefer detailed explanations, while others might seek clarity and brevity.
Moreover, a pun that amuses one person may offend another. 
Future studies can consider designing more appropriate evaluation metrics.

\section*{Ethics Statement}
\label{sec:ethic}
We acknowledge that all authors are informed about and adhere to the ACL Code of Ethics and the Code of Conduct.

\paragraph{Use of Human Annotations}
Our institution recruited annotators to implement the annotations of pun evaluation. 
We ensure the privacy rights of the annotators are respected during the annotation process.
The annotators receive compensation exceeding the local minimum wage and have consented to the use of pun data generated by them for research purposes. Appendix~\ref{appendix:crowd-sourcing} provides further details on the annotations.

\paragraph{Risks}
The pun datasets in our experiment are sourced from publicly available sources. However, we cannot guarantee that they are devoid of socially harmful or toxic language. 
Furthermore, evaluating the data quality of pun explanation and generation is based on common sense, which can vary among individuals from diverse backgrounds.
We use ChatGPT to correct grammatical errors in this paper.

\bibliography{anthology,custom}

\clearpage
\appendix
\label{sec:appendix}
\section{Crowd-sourcing}
\label{appendix:crowd-sourcing}
\begin{figure*}[ht]
    \centering
    \small
    \includegraphics[width=0.9\linewidth]{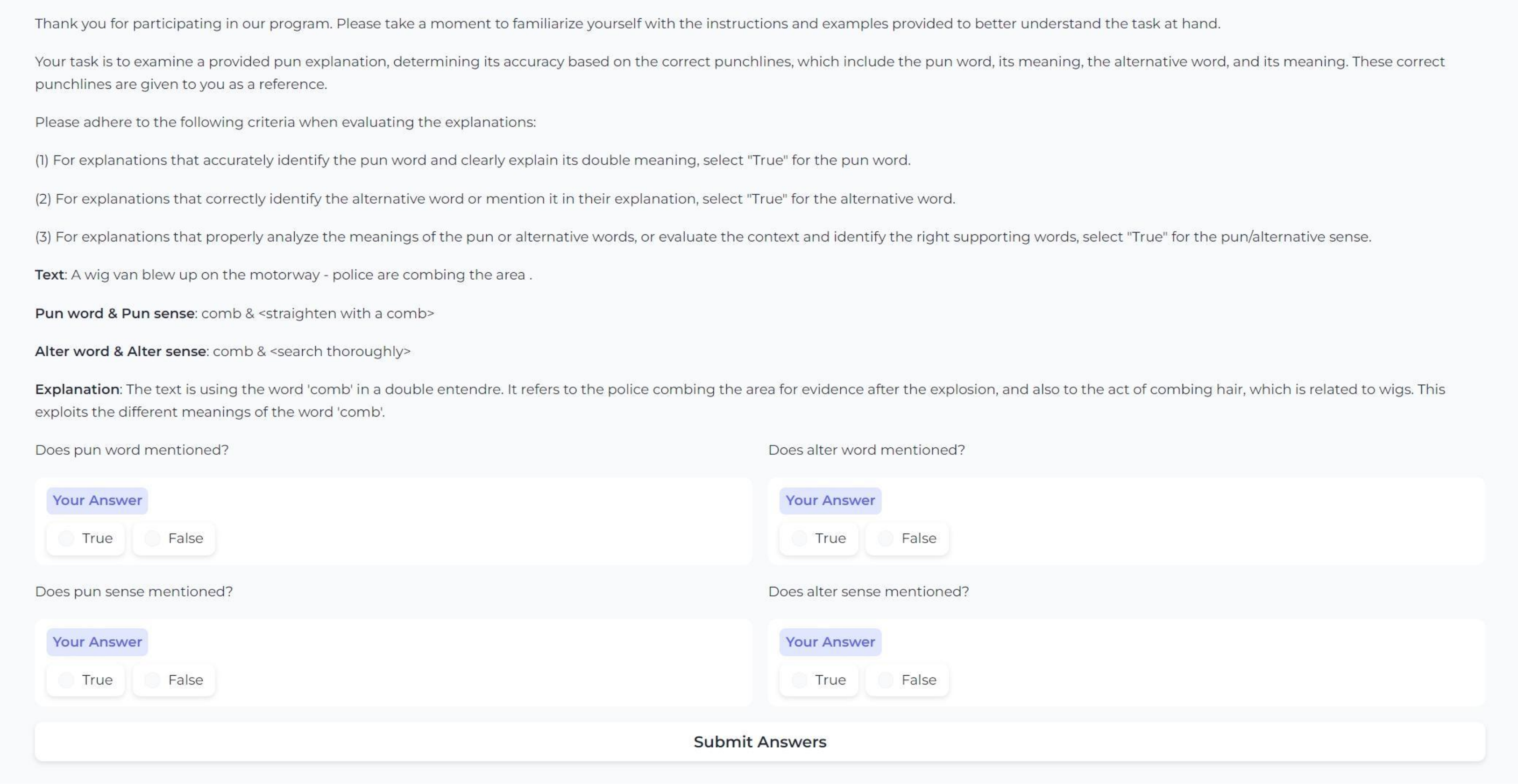}
    \caption{The screenshot of punchline check annotation.
    }
    \label{fig:punchline_interface}
\end{figure*}

\begin{figure*}[ht]
    \centering
    \small
    \includegraphics[width=0.9\linewidth]{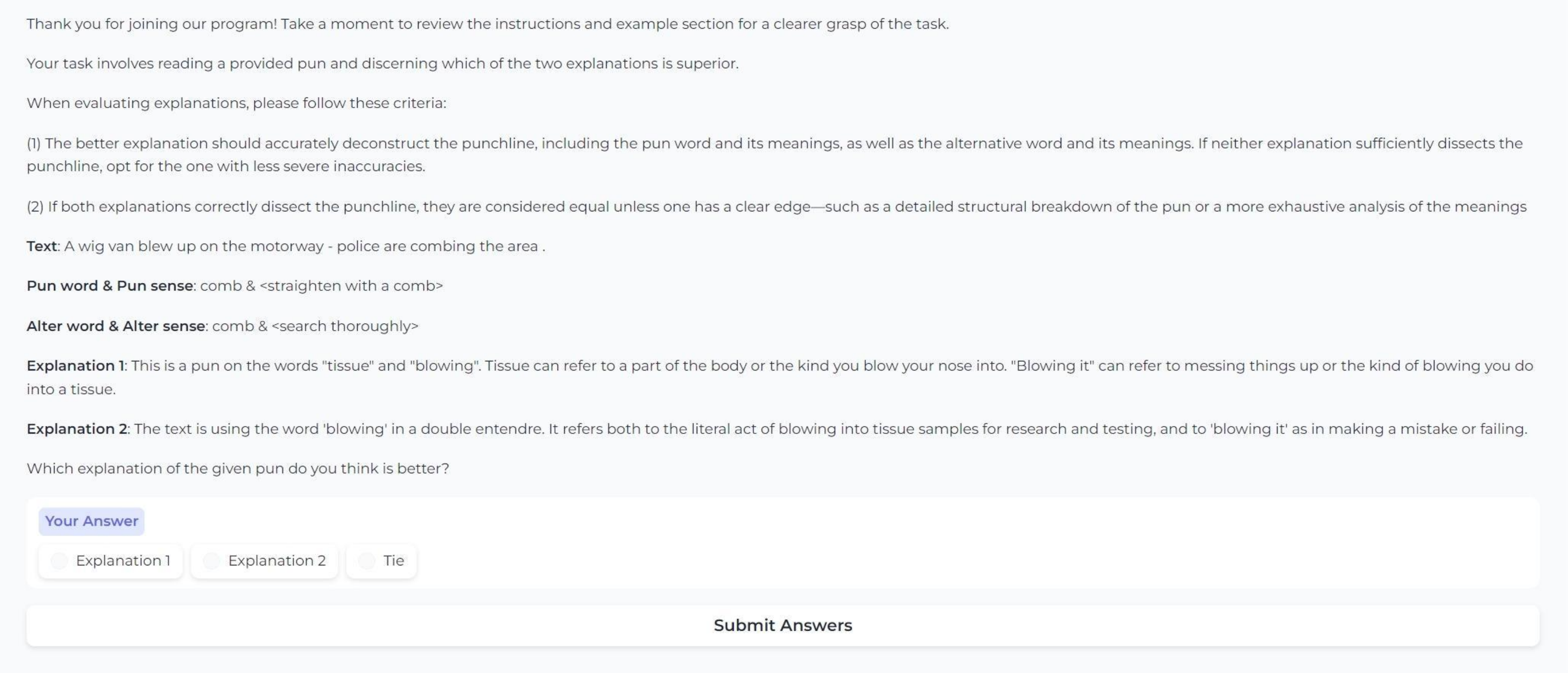}
    \caption{The screenshot of pairwise comparison annotation.
    }
    \label{fig:paircmp_interface}
\end{figure*}

\begin{figure*}[ht]
    \centering
    \small
    \includegraphics[width=0.9\linewidth]{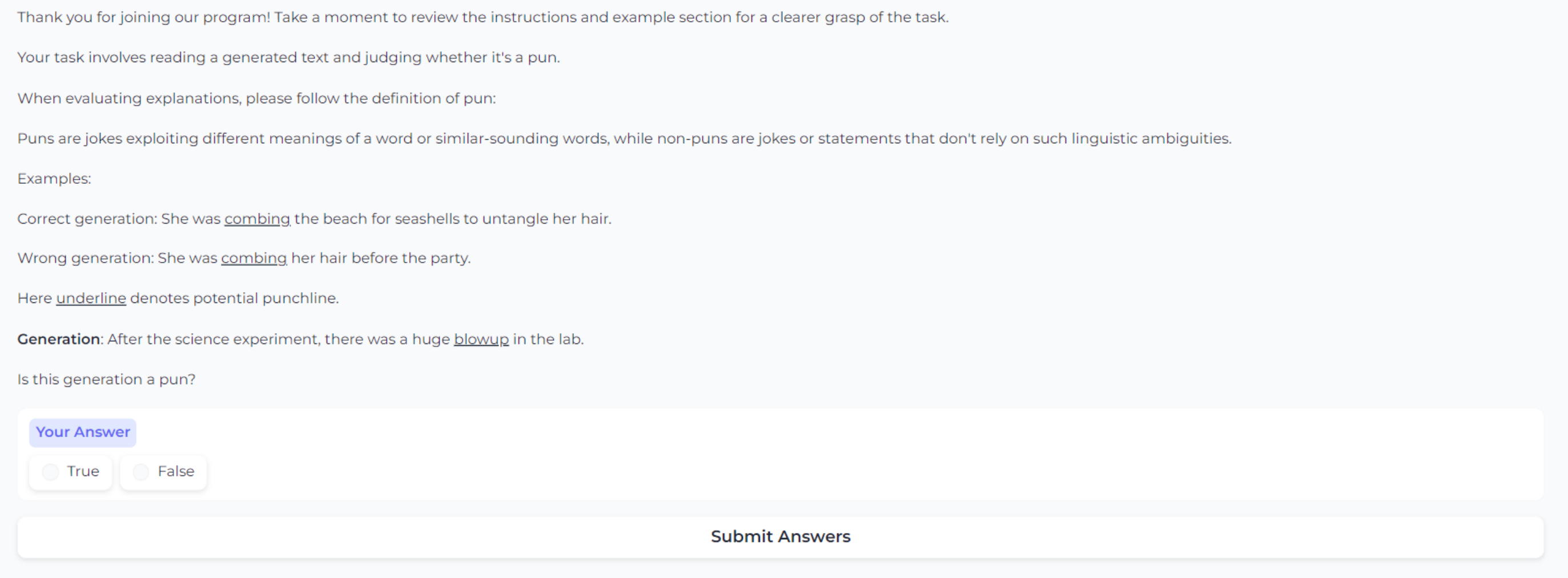}
    \caption{The screenshot of generation success annotation.
    }
    \label{fig:gen_interface}
\end{figure*}

We have recruited a team of three undergraduates who majored in English.
They are very familiar with puns and are specifically trained for our evaluation work.
We pay each of them \$9/h, exceeding the local minimum wage. The screenshots of the instructions and annotation interface are shown in Figure~\ref{fig:punchline_interface},~\ref{fig:paircmp_interface},~\ref{fig:gen_interface}.

\section{Details of Prompts}
\label{appendix:prompt}
\subsection{Prompt for Pun Recognition and Explanation}
\label{appendix:promptrec}
When recognizing different types of puns, we will provide accordingly 6 examples.
The explanations used in experiments are collected from enhanced prompts for best performance.
We set the temperature parameter to 0 to minimize the impact of the model's inherent randomness on recognition and explanation tasks.
\begin{table}[H]
\footnotesize
  \centering
    \begin{tabularx}{\linewidth}{X}
    \toprule
    \rowcolor[gray]{0.95}\multicolumn{1}{c}{\textbf{Pun Recognition and Explanation}} \\
    \midrule
     \makecell[{{p{\linewidth}}}]{
     \vspace{-2.5mm}
    \color{gray}{/* \textit{Definition} */}\\
    Puns are a form of wordplay exploiting different meanings \\ of a word or similar-sounding words, while non-puns are \\ jokes or statements that don't rely on such linguistic ambiguities. \\ 
    \color{gray}{/* \textit{Instruction} */} \\
    Determine whether the given Text is a \textbf{pun/non-pun.} \textcolor{red}{Give}\\ \textcolor{red}{your reasons first, then make your final decision clearly.} \\ You should either say "The given text is a pun" or say "The\\ given text is a non-pun".  You must output the current status\\ in a parsable JSON format. An example output looks like \\ \{\textcolor{red}{"Reason": "XXX"}, "Choice": "The given text is a XXX"\} \\  
    \color{gray}{/* \textit{6 Examples} */} \\
    Text: Driving on so many turnpikes was taking its toll.\\
    Output:\{\textcolor{red}{"Reason": "The text is using the word 'toll' in a do-}\\\textcolor{red}{uble entendre. It refers both to the physical tolls paid on tu-}\\\textcolor{red}{rnpikes and to 'taking its toll' as in having a negative effect} \\ \textcolor{red}{or cost."}, "Choice": "The Given test is a pun."\}\\
    Text: Nothing ventured, nothing gained.\\
    Output:\{\textcolor{red}{"Reason": "The given text is a proverb that expres-}\\ \textcolor{red}{ses a general truth or piece of advice and does not exploit}\\ \textcolor{red}{different meanings of a word or similar-sounding words.},\\ "Choice": "The given text is a non-pun."\}\\
    ......\\ 
    \color{gray}{/* \textit{Test Data} /}\\
    Text: I wanted to have dinner at a native American-themed \\ restaurant, but I didn't have reservations.\\
    Output:}\\
    \bottomrule
    \end{tabularx}
  \caption{Prompt for pun recognition and explanation.
  \textcolor{red}{Red Text} denotes the Chain of Thought (CoT) module.
  We will select a single bias indicated by \textbf{bold text} at a time. 
  }
  \label{tab:prompt_recog}
\end{table}

\subsection{Prompt for Pun Generation}
\label{appendix:promptgen}
In pun generation tasks, we will provide 3 examples in the prompt and test the effect of contextual words on the final generation's quality. 
Here, the temperature parameter is set to 0.7, which strikes a balance between stimulating the model's creativity and preventing it from going off the rails.
\begin{table}[h!]
\footnotesize
  \centering
    \begin{tabularx}{\linewidth}{X}
    \toprule
    \rowcolor[gray]{0.95}\multicolumn{1}{c}{\textbf{Pun Generation}} \\
    \midrule
    \makecell[{{p{\linewidth}}}]{
    \vspace{-2.5mm}
    \color{gray}{/* \textit{Definition} */} \\
    Puns are a form of wordplay exploiting different meanings of a word or similar-sounding words, while non-puns are jokes or statements that don't rely on such linguistic ambiguities.\\
    \color{gray}{/* \textit{Instruction} */} \\
    Below is a keyword, two of its meanings \textcolor{red}{and a set of contextual words.}
    Please generate a pun sentence with a punchline on the keyword that conveys both given meanings simultaneously and using all the contextual words. 
    Except for the keyword, the pun sentence must not utilize any words from either of the two meanings. Besides, once a keyword is used, it's strictly prohibited to use it again in the latter half of the sentence.
    You must output the current status in a parsable JSON format. An example output looks like:
    \{"Sentence": "XXX"\}\\
    \color{gray}{/* \textit{3 Examples */}} \\
    Keyword: toll\\
    Meaning 1: toll <a fee levied for the use of roads or bridges (used for maintenance)>\\
    Meaning 2: toll <value measured by what must be given or done or undergone to obtain something>\\
    \textcolor{red}{Contextual Words: Driving, many, turnpikes, taking its toll}\\
    Output:\\
    \{"Sentence": "\{"Driving on so many turnpikes was taking its toll."\}"\}\\
    ......\\
    \color{gray}{/* \textit{Test Data} */} \\
    Keyword: bore\\
    Meaning 1: <Make a hole, especially with a pointed power or hand tool>\\
    Meaning 2: <A carpenter sat on his drill and was bored to tears.>\\
    \textcolor{red}{Contextual Words: carpenter, sat, drill, bored to tears}\\
    Output:\\
    }\\
    \bottomrule
    \end{tabularx}
  \caption{Prompt for pun generation.
  {\color{red}{Red texts}} denotes the addition of contextual words.
  }
  \label{tab:prompt_pungen}
\end{table}

\subsection{Prompt for Non-pun Generation}
\label{appendix:promptgennon}
We use GPT3.5-Turbo to generate non-puns as lower-bound references for the evaluation metric.
This task is relatively simple so we don't provide examples.
The prompt is presented in Table~\ref{tab:prompt_nonpungen}.

\subsection{Prompt for Pairwise Comparison}
\label{appendix:paircmp}
During the preliminary experiments of pairwise comparison, we provide GPT-4 with three examples for reference.
However, we later noticed that the model's performance is similar with both 0-shot and 3-shot settings.
Considering that not providing examples could significantly save on token usage, we ultimately opt for the 0-shot approach.
The prompt is placed in Table~\ref{tab:prompt_pairwise}.
\begin{table}[h!]
\footnotesize
  \centering
    \begin{tabularx}{\linewidth}{X}
    \toprule
    \rowcolor[gray]{0.95}\multicolumn{1}{c}{\textbf{Non-pun Generation}} \\
    \midrule
     \makecell[{{p{\linewidth}}}]{
     \vspace{-2.5mm}
    \color{gray}{/* \textit{Definition} */} \\
    Puns are a form of wordplay exploiting different meanings of a word or similar-sounding words, while non-puns are jokes or statements that don't rely on such linguistic ambiguities.\\
    \color{gray}{/* \textit{Instruction} */} \\
    Below is a keyword and one of its meanings. Please generate a non-pun sentence with the keyword that conveys the given meaning. 
    You must output the current status in a parsable JSON format. 
    An example output looks like:
    \{"Sentence": "XXX"\}\\
    \color{gray}{/* \textit{Test Data} */} \\
    Keyword: thick\\
    Meaning: <having a short and solid form or stature>\\
    Output:\\
    }\\
    \bottomrule
    \end{tabularx}
  \caption{Prompt for non-pun generation}
  \label{tab:prompt_nonpungen}
\end{table}

\begin{table}[h!]
\footnotesize
  \centering
    \begin{tabularx}{\linewidth}{X}
    \toprule
    \rowcolor[gray]{0.95}
    \multicolumn{1}{c}{\textbf{Pairwise Comparison}} \\
    \midrule
    \makecell[{{p{\linewidth}}}]{
    \vspace{-2.5mm}
    \color{gray}{/* \textit{Definition} */} \\
    Puns are a form of wordplay exploiting different meanings of a word or similar-sounding words.\\
    \color{gray}{/* \textit{Instruction} */} \\
    Below is a pun text, gold meanings of pun, and two corresponding explanations. Please carefully judge which explanation is of better quality. A good explanation should point out the correct pun word and analyze the multiple meanings of the pun or similar-sounding words in detail appropriately while avoiding unnecessary or incorrect interpretations. You must choose from one of the three answers: "Explanation 1 is much better", "Explanation 2 is much better", "I'm not sure which would be better.". You must output the current status in a parsable JSON format. An example output looks like: \{"Choice": "XXX"\} \\ 
    \color{gray}{/* \textit{Text Data} */} \\
    Pun Text: Have another soft drink, Tom coaxed.\\
    Gold Meanings of Pun: \\
    1. coax < influence or urge by gentle urging, caressing, or flattering > \\
    2. coke < Coca Cola is a trademarked cola >\\
    Explanation 1: This is a pun on how "coaxed" sounds like "Coke" which is a brand of soft drink.\\
    Explanation 2: The text plays on the double meaning of the word 'coaxed'. "Coaxed" can mean persuading someone to do something, but it can also refer to mixing or stirring a drink. This creates a humorous double meaning.\\
    Output:\\
    }\\
    \bottomrule
    \end{tabularx}
  \caption{Prompt for pairwise comparison.}
  \label{tab:prompt_pairwise}
\end{table}

\subsection{Prompt for Finding Synonyms}
For assessing ambiguity, distinctiveness, surprise, and unusualness, synonyms play a crucial role in the calculations, as detailed in Appendix~\ref{appendix:adsuImpl}. 
So we design a prompt to find synonyms for both the pun words and alternative words in hom-puns.
We use GPT-4 to complete this work.

\begin{table}[h!]
\footnotesize
  \centering
    \begin{tabularx}{\linewidth}{X}
    \toprule
    \rowcolor[gray]{0.95}
    \multicolumn{1}{c}{\textbf{Finding Synonyms}} \\
    \midrule
    \makecell[{{p{\linewidth}}}]{
    \vspace{-2.5mm}
    \color{gray}{/* \textit{Instruction} */} \\
    Below is a pun text, one keyword, and its two meanings. The keyword is the pun in the text, which can be interpreted in two meanings. Please find two different synonyms for the keyword, each corresponding to one of the meanings. The synonyms should be able to replace the keyword in the text seamlessly to remove ambiguity, while ideally being a simple word. You must output the current status in a parsable JSON format. An example output looks like:
    \{'Synonym 1 for Meaning 1': 'XXX', 'Synonym 2 for Meaning 2': 'XXX'\}    \\
    \color{gray}{/* \textit{6 Examples} */} \\
    Text: Driving on so many turnpikes was taking its toll.\\
    Keyword: toll\\
    Meaning 1: < a fee levied for the use of roads or bridges (used for maintenance) >\\
    Meaning 2: < value measured by what must be given or done or undergone to obtain something >\\
    Output:\\
    \{"Synonym 1 for Meaning 1": "fee", "Synonym 2 for Meaning 2": "impact"\}\\
    ......\\ 
    \color{gray}{/* \textit{Test Data} */} \\
     Text: A boy told his parents he wanted to raise goats for a living, but he was only kidding. \\
    Keyword: kid\\
    Meaning 1: < tell false information to for fun >\\
    Meaning 2: < young goat >\\
    Output:\\
    }\\
    \bottomrule
    \end{tabularx}
  \caption{Prompt for finding synonyms.}
  \label{tab:prompt_synoym}
\end{table}

\section{Details of A, D, S, and U Metrics}
\label{appendix:adsu}
\subsection{Formulas}

\paragraph{Ambiguity \& Distinctiveness~\cite{kao2016computational}} Ambiguity measures the extent to which the sentence supports both pun sense and alternative sense. It's quantified by the entropy of $P(m|\mathbf{w})$, where $m$ is either the pun word $w_p$ or the alternative word $w_a$.

\begin{equation*}
P(m|\mathbf{w}) = \sum_{\mathbf{f}} \left( P(m)P(\mathbf{f})\prod_{i} P(w_i|m, f_i) \right)
\end{equation*}

Distinctiveness is indicative of how distinctive the meanings $m1$ ($w_p$) and $m2$ ($w_a$) are, based on the supporting subsets of words in the sentence and it's calculated by KL divergence.

\begin{equation*}
D_{KL}(F1||F2) + D_{KL}(F2||F1)
\end{equation*}

Variables:
\begin{itemize}
  \item $m$: Pun word or alternative word.
  \item $\mathbf{w}$: Context.
  \item $\mathbf{f}$: Indicate whether a certain word is related to the topic.
  \item $F1$, $F2$: Distributions of focus sets given sentence topics $m1$ and $m2$, respectively.
  \item $D_{KL}$: Symmetrized Kullback-Leibler divergence score representing the distinctiveness between $F1$ and $F2$.
\end{itemize}

\paragraph{Surprise \& Unusualness~\cite{he-etal-2019-pun}}
Surprise in puns arises from the unexpected presence of the pun word over an anticipated one within a sentence, generating humor. It's quantified by $S_{ratio}$.

\begin{align*}
S(c) &= -\log \left( \frac{p(w_p | c)}{p(w_a | c)} \right) \\
S_{\text{local}} &= S(x_{p-d:p-1}, x_{p+1:p+d}), \nonumber \\
S_{\text{global}} &= S(x_{1:p-1}, x_{p+1:n}), \nonumber \\
S_{\text{ratio}} &= \begin{cases} 
-1, & \text{if } S_{\text{local}} < 0 \text{ or } S_{\text{global}} < 0, \\
\frac{S_{\text{local}}}{S_{\text{global}}}, & \text{otherwise}. 
\end{cases} \nonumber
\end{align*}

Unusualness attends to the pun's anomalous nature, and it's quantified by:
\[
\text{Unusualness} \overset{\mathrm{def}}{=} -\frac{1}{n} \log \left( \frac{p(x_1, \ldots, x_n)}{\prod_{i=1}^{n} p(x_i)} \right)
\]

Variables are as follows:
\begin{itemize}
  \item \( w_p \): Pun word.
  \item \( w_a \): Expected alternative word.
  \item \( c \): Context.
  \item \( S_{\text{local}} \): Local surprisal.
  \item \( S_{\text{global}} \): Global surprisal.
\end{itemize}

\subsection{Implementation}
\label{appendix:adsuImpl}
Similar to previous papers, we apply a SkipGram model~\cite{mikolov2013efficient} to evaluate \textbf{A}mbiguity and \textbf{D}istinctiveness, and use an LM pre-trained on WikiText~\cite{merity2016pointer} to evaluate \textbf{S}urprise.
 Given that these models had limited vocabularies, we only calculate metrics for generations with words within the model's lexicon. Additionally, we exclude the top 2\% of extreme values in \textbf{S} calculations to prevent distortion of results due to near-zero denominators.

These metrics based on word probability modeling require the pun words and alternative words to differ, which is not a problem for het-puns but poses a challenge for hom-puns.
To address this issue, we adopt the idea of \citet{tian-etal-2022-unified}.
This involves finding synonyms for both $w_p$ and $w_a$ according to the double meanings in hom-puns and then using these synonyms as substitutes for calculating the metrics. The prompt for synonym generation can be referenced in Table~\ref{tab:prompt_synoym}.

\section{Case Study of LLM-Generated Explanations}
\subsection{Text Structure}
\label{appendix:structure}
When interpreting puns, LLMs typically start by identifying the $w_p$ and $w_a$ and then analyze the dual meanings in the surrounding context.
This clear main-and-subordinate structure is not very common in human explanations, which makes them less intuitive than those by models, as shown in Table~\ref{tab:explcase}.
\begin{table}[ht]
\footnotesize
  \centering
    \begin{tabularx}{\linewidth}{X}
    \toprule
    \rowcolor[gray]{0.95}
    \multicolumn{1}{c}{\textbf{Pun Explanation Samples}} \\
    \midrule
    \makecell[{{p{\linewidth}}}]{
    \vspace{-2.5mm}
    \color{gray}{/* \textit{Pun Text} */}\\
    Actors are cast through their good and bad stages.\\
    \color{gray}{/* \textit{Human Explanation} */} \\
    This sentence is a pun. It's saying that actors are cast through good and bad acting. The word used is stages instead of acting. A stage is a raised platform that actors act on.\\
    \color{gray}{/* \textit{LLM Explanation} */} \\
   The text \underline{plays on the double meaning of the word 'stages'}. It refers to both the phases of an actor's career (good and bad times) and the physical stages (platforms) in theaters where actors perform.\\
    }\\ 
    \midrule
    \makecell[{{p{\linewidth}}}]{
    \vspace{-2.5mm}
    \color{gray}{/* \textit{Pun Text} */}\\
    Is that a pure rump roast? Nothing butt.\\
    \color{gray}{/* \textit{Human Explanation} */} \\
    This is a joke on the fact that rump roasts come from the butt of the animal, and the speaker answering the question here is confirming that the product is nothing but pure rump roast roast.\\
    \color{gray}{/* \textit{LLM Explanation} */} \\
   The text \underline{plays on the homophones 'butt' and 'but'}, using 'butt' in the context of a rump roast (which refers to the rear or buttocks of the animal) and 'but' as a conjunction typically used to introduce a phrase or clause contrasting with what has already been mentioned. This creates a humorous double meaning.\\
    }\\ 
    \bottomrule
    \end{tabularx}
  \caption{Comparison between human pun explanations and LLM-generated pun explanations.
  We \underline{underline} the segments that explicitly point out the $w_p$ and $w_a$.}
  \label{tab:explcase}
\end{table}

% Text: Actors are cast through their good and bad stages.
% Human expl: This sentence is a pun. It's saying that actors are cast through good and bad acting. The word used is stages instead of acting. A stage is a raised platform that actors act on.
% GPT4-Turbo expl: The text plays on the double meaning of the word 'stages.' It refers to both the phases of an actor's career (good and bad times) and the physical stages (platforms) in theaters where actors perform.

% Text: Is that a pure rump roast? Nothing butt.
% Human expl: This is a joke on the fact that rump roasts come from the butt of the animal, and the speaker answering the question here is confirming that the product is nothing but pure rump roast roast
% GPT-4-Turbo expl: The text plays on the homophones 'butt' and 'but', using 'butt' in the context of a rump roast (which refers to the rear or buttocks of the animal) and 'but' as a conjunction typically used to introduce a phrase or clause contrasting with what has already been mentioned. This creates a humorous double meaning.

\subsection{Error Types}
\label{appendix:explerror}
In our analysis, we've identified recurring error patterns in the output of LLMs and will examine these trends in the forthcoming case studies. 
All results below are acquired from GPT-3.5's result, which we find struck a balance between generation quality and noticeability of erroneous patterns.

\paragraph{Misclassify Pun as Non-pun}
For certain common puns, LLM tends to classify them as proverbs instead of puns. This shows LLM's comprehension of puns is superimposed by its understanding of certain common expressions.
\begin{table}[ht]
\footnotesize
  \centering
    \begin{tabularx}{\linewidth}{X}
    \toprule
    \rowcolor[gray]{0.95}
    \multicolumn{1}{c}{\textbf{Misclassify Pun as Non-pun}} \\
    \midrule
    \makecell[{{p{\linewidth}}}]{
    \vspace{-2.5mm}
    \color{gray}{/* \textit{Pun Text} */}\\
    He who lends to the poor gets interest from god.\\
    \color{gray}{/* \textit{Right explanation of Human} */}\\
   This is a pun about earning interest, as in interest rates on money and fascination, from god. Helping a poor person is considered a good gesture in the eyes of religion.\\
    \color{gray}{/* \textit{Wrong Explanation of LLM} */} \\
   The text is a proverb conveying a moral lesson and does not exploit different meanings of a word or similar-sounding words for humorous effect.\\
    }\\ 
    \bottomrule
    \end{tabularx}
  \caption{LLM misclassify pun as non-pun in explanation}
  \label{tab:nonpunerr}
\end{table}

\paragraph{Incorrect Pun Word Identification}
For certain het-puns, LLM can't correctly identify the pun word because the relationship between the pun word and alternative words is a bit far. Like in this example, we need to associate "slow" with "sloe" using phonetic similarity and then associate "sloe" with "blackthorn" by semantic similarity. These types of puns are challenging for current LLMs.
\begin{table}[ht]
\footnotesize
  \centering
    \begin{tabularx}{\linewidth}{X}
    \toprule
    \rowcolor[gray]{0.95}
    \multicolumn{1}{c}{\textbf{Incorrect Pun Word Identification}} \\
    \midrule
    \makecell[{{p{\linewidth}}}]{
    \vspace{-2.5mm}
    \color{gray}{/* \textit{Pun Text} */}\\
    'This looks like the fruit of the blackthorn', said Tom slowly.\\
    \color{gray}{/* \textit{Right explanation of Human} */} \\
    This is a play on words. The word "slowly" means at a slow speed but a "sloe" is the fruit of a blackthorn..\\
    \color{gray}{/* \textit{Wrong Explanation of LLM} */} \\
    The text plays on the double meaning of 'blackthorn', which refers to both the fruit of the blackthorn plant and the color black. This creates a humorous double meaning.\\
    }\\ 
    \bottomrule
    \end{tabularx}
    \caption{LLM incorrectly pinpoint pun word in explanation}
  \label{appendix:caseb}
\end{table}

\paragraph{Incorrect Alternative Word Identification}
In some het puns, LLM's explanation shows a strange pattern: it can't identify the correct alternative word, but it can give out the right alternative meaning. Like in this example, it doesn't pinpoint the pun word: "canvass" but the explanation includes its meaning: "the act of going door-to-door to promote a product or service". For these cases, we think LLM truly understands both meanings of the sentence. However, its inability to wield phonetic similarity leads the failure in alternative word identification.
% \newpage
\begin{table}[ht]
\footnotesize
  \centering
    \begin{tabularx}{\linewidth}{X}
    \toprule
    \rowcolor[gray]{0.95}
    \multicolumn{1}{c}{\textbf{Incorrect Alternative Word Identification}} \\
    \midrule
    \makecell[{{p{\linewidth}}}]{
    \vspace{-2.5mm}
    \color{gray}{/* \textit{Pun Text} */}\\
    In his latest promotional campaign, the manager of City Tent and Awning told his crew to canvas the entire area.\\
    \color{gray}{/* \textit{Right explanation of Human} */} \\
   "Canvas" can both mean a noun that refers to a type of rough, durable cloth or fabric commonly used in tents and outdoor coverings such as awnings, and a verb that means searching or covering a given area, usually in the context of personal outreach for a given goal or mission. The play on words here is that the manager of a business selling tents and awnings, commonly made of canvas, is telling his employees to help promote the business by *canvassing* the area to that end.\\
    \color{gray}{/* \textit{Wrong Explanation of LLM} */} \\
   The text plays on the double meaning of 'canvas'. In the context of a promotional campaign, 'canvas' refers to the act of going door-to-door to promote a product or service. However, 'canvas' is also a material used for making tents and awnings, which relates to the manager's business. This creates a humorous double meaning.\\
    }\\ 
    \bottomrule
    \end{tabularx}
    \caption{LLM incorrectly pinpoint alternative word in explanation}
  \label{appendix:casec}
\end{table}

\paragraph{Misinterpret Het as Hom}
A frequently appearing pattern in het-puns explanations. These explanations include expressions such as: "use the similar sounding of", and "plays on the homophones of". Demonstrating that LLM doesn't have a firm grasp of the definition of het-pun and hom-pun.
\begin{table}[htbp]
\footnotesize
  \centering
    \begin{tabularx}{\linewidth}{X}
    \toprule
    \rowcolor[gray]{0.95}
    \multicolumn{1}{c}{\textbf{Misinterpret Het as Hom}} \\
    \midrule
    \makecell[{{p{\linewidth}}}]{
    \vspace{-2.5mm}
    \color{gray}{/* \textit{Pun Text} */}\\
    "I've been listening to the Brandenburg Concertos", Tom barked\\
    \color{gray}{/* \textit{Right explanation of Human} */} \\
    'Bark' sounds like Bach, who is the composer of Brandenburg Concertos. Using 'bark' creates a humorous effect\\
    \color{gray}{/* \textit{Wrong Explanation of LLM} */} \\
    The text plays on the homophones 'barked' and 'barked', using 'barked' in the context of a dog making a sound and 'barked' as in speaking loudly and aggressively. This creates a humorous double meaning.\\
    }\\ 
    \bottomrule
    \end{tabularx}
    \caption{LLM misinterpret het-pun as hom-pun in explanation}
  \label{appendix:cased}
\end{table}

\paragraph{Lack of Meaning Analysis}
For certain common puns, especially het puns, LLM will only pinpoint the pun word and alternative words without an explanation of their respective meanings.
\begin{table}[ht]
\footnotesize
  \centering
    \begin{tabularx}{\linewidth}{X}
    \toprule
    \rowcolor[gray]{0.95}
    \multicolumn{1}{c}{\textbf{Lack of Meaning Analysis}} \\
    \midrule
    \makecell[{{p{\linewidth}}}]{
    \vspace{-2.5mm}
    \color{gray}{/* \textit{Pun Text} */}\\
    All students in Helsinki go to Finnish school.\\
    \color{gray}{/* \textit{Right explanation of Human} */}\\
   This is a pun on 'finishing school' which is a private school where girls are prepared for entry into fashionable society, but 'Finnish' is used in its place because it sounds slightly similar and since Helsinki is the capital of Finland, any school there is a Finnish school.\\
    \color{gray}{/* \textit{Wrong Explanation of LLM} */} \\
   The text plays on the similarity in sound between 'Finnish' and 'finish', creating a humorous double meaning. It exploits the similar-sounding words for a humorous effect.\\
    }\\ 
    \bottomrule
    \end{tabularx}
  \caption{LLM doesn't explain meanings of pun/alternative words in explanation}
  \label{tab:nonpunerror}
\end{table}

\paragraph{Fabricating Non-existent Meanings}
After LLM wrongly identifies a pun word or alternative word, they will continue to explain the pun based on the previous incorrect conclusion. Just like the following example shows, after "plays on the double meaning of splice", LLM generated the nonexistent meaning of "splice" as "adding toppings to a pizza" to fit in the explanation.
\begin{table}[ht]
\footnotesize
  \centering
    \begin{tabularx}{\linewidth}{X}
    \toprule
    \rowcolor[gray]{0.95}
    \multicolumn{1}{c}{\textbf{Fabricating Non-existent Meanings}} \\
    \midrule
    \makecell[{{p{\linewidth}}}]{
    \vspace{-2.5mm}
    \color{gray}{/* \textit{Pun Text} */}\\
    Did you hear about the film editor at Hollywood who had a part-time job at a pizza parlor? He used to ask customers whether they wanted one splice or two.\\
    \color{gray}{/* \textit{Right explanation of Human} */} \\
    "Splice" sounds like "slice" and "slice" is a word often associated with pizza. "Splice" is a verb that means to join or connect. In film, a splice is when a film has to be taped together because of breakage. This is often seen during the editing process. The joke is about a film editor's part-time job at a pizza parlor.\\
    \color{gray}{/* \textit{Wrong Explanation of LLM} */} \\
    The text plays on the double meaning of 'splice', which can refer to both editing film and adding toppings to a pizza. This creates a humorous play on words.\\
    }\\ 
    \bottomrule
    \end{tabularx}
  \caption{LLM makes up non-existent meanings in explanation}.
  \label{tab:hallucinationerror}
\end{table}

\newpage

\end{document}